\documentclass[journal]{IEEEtran}

\usepackage{csvsimple}
\usepackage{booktabs}
\usepackage{array}
\usepackage{stfloats}
\usepackage{upgreek}
\usepackage{saurav_macros}
\usepackage{mathtools}
\newcommand*{\scost}[1]{c_s(#1)\xspace} 
\newcommand*{\dcost}[1]{c_d(#1)\xspace} 
\newcommand*{\sdemand}[1]{q_s(#1)\xspace} 
\newcommand*{\ddemand}[1]{q_d(#1)\xspace} 

\newcommand*{\svk}[2][k]{s_{#2}^{#1}} 
\newcommand*{\ddk}[2][k]{d_{#2}^{#1}} 
\newcommand*{\zk}[2][k]{z_{#2}^{#1}} 

\usepackage[caption=false,font=footnotesize]{subfig}

\usepackage{fancyhdr}
\usepackage{pgfplots}
\pgfplotsset{width=0.49\textwidth,compat=1.6}
\usetikzlibrary{pgfplots.statistics}

\usepackage{color}

\usepackage[normalem]{ulem}
\fancypagestyle{specialfooter}{%
  \fancyhf{}
  
  \fancyfoot[C]{\scriptsize The paper will appear in the IEEE Transactions on Robotics. \textcopyright 2023 IEEE.  Personal use of this material is permitted.  Permission from IEEE must be obtained for all other uses, in any current or future media, including reprinting/republishing this material for advertising or promotional purposes, creating new collective works, for resale or redistribution to servers or lists, or reuse of any copyrighted component of this work in other works.}
}

\begin{document}
\author{Saurav Agarwal and Srinivas Akella%
	\thanks{This work was supported in part by NSF awards IIS-1547175, IIP-1919233 and IIP-1439695, a UNC IPG award, and DARPA HR00111820055.}
	\thanks{This paper has supplementary downloadable video provided by the authors.
	The video provides a brief overview of the line coverage problem, illustrates the primary algorithm, and showcases simulations and experiments.}
	\thanks{The authors are with the Department of Computer Science, University of North Carolina at Charlotte, NC 28223, USA. Corresponding author: Saurav Agarwal. 
	E-mail addresses: {\tt\footnotesize \{sagarw10,sakella\}@charlotte.edu}}%
}
\title{Line Coverage with Multiple Robots:\\Algorithms and Experiments}
\maketitle
\thispagestyle{specialfooter}

\begin{abstract}
The line coverage problem involves finding efficient routes for the coverage of linear features by one or more resource-constrained robots. Linear features model environments like road networks, power lines, and oil and gas pipelines. Two modes of travel are defined for robots: servicing and deadheading. A robot services a feature if it performs task-specific actions, such as taking images, as it traverses the feature; otherwise, it is deadheading. Traversing the environment incurs costs (e.g., travel time) and demands on resources (e.g., battery life). Servicing and deadheading can have different cost and demand functions, which can be direction-dependent. The environment is modeled as a graph, and an integer linear program is provided. As the problem is NP-hard, we design a fast and efficient heuristic algorithm, Merge-Embed-Merge (MEM). Exploiting the constructive property of the MEM algorithm, algorithms for line coverage of large graphs with multiple depots are developed. Furthermore, turning costs and nonholonomic constraints are efficiently incorporated into the algorithm. The algorithms are benchmarked on road networks and demonstrated in experiments with aerial robots.
\end{abstract}

\section{Introduction}
\label{sc:intro}
\begin{figure*}[htbp]
	\centering
	\subfloat[Input Road Network]{%
	\includegraphics[height=0.23\textheight]{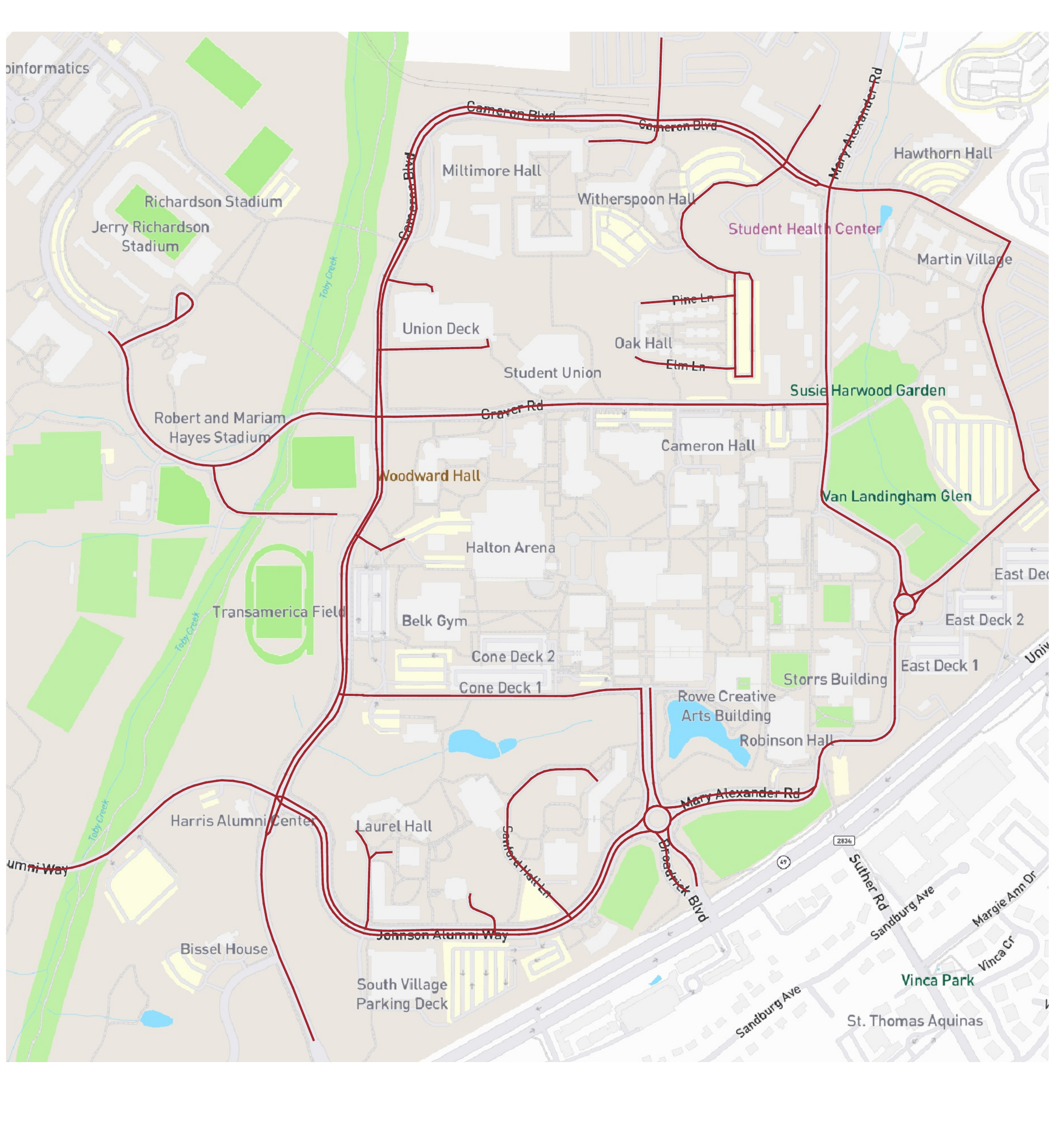}}
	\hfill
	\subfloat[Line Coverage Routes]{%
	\includegraphics[height=0.23\textheight]{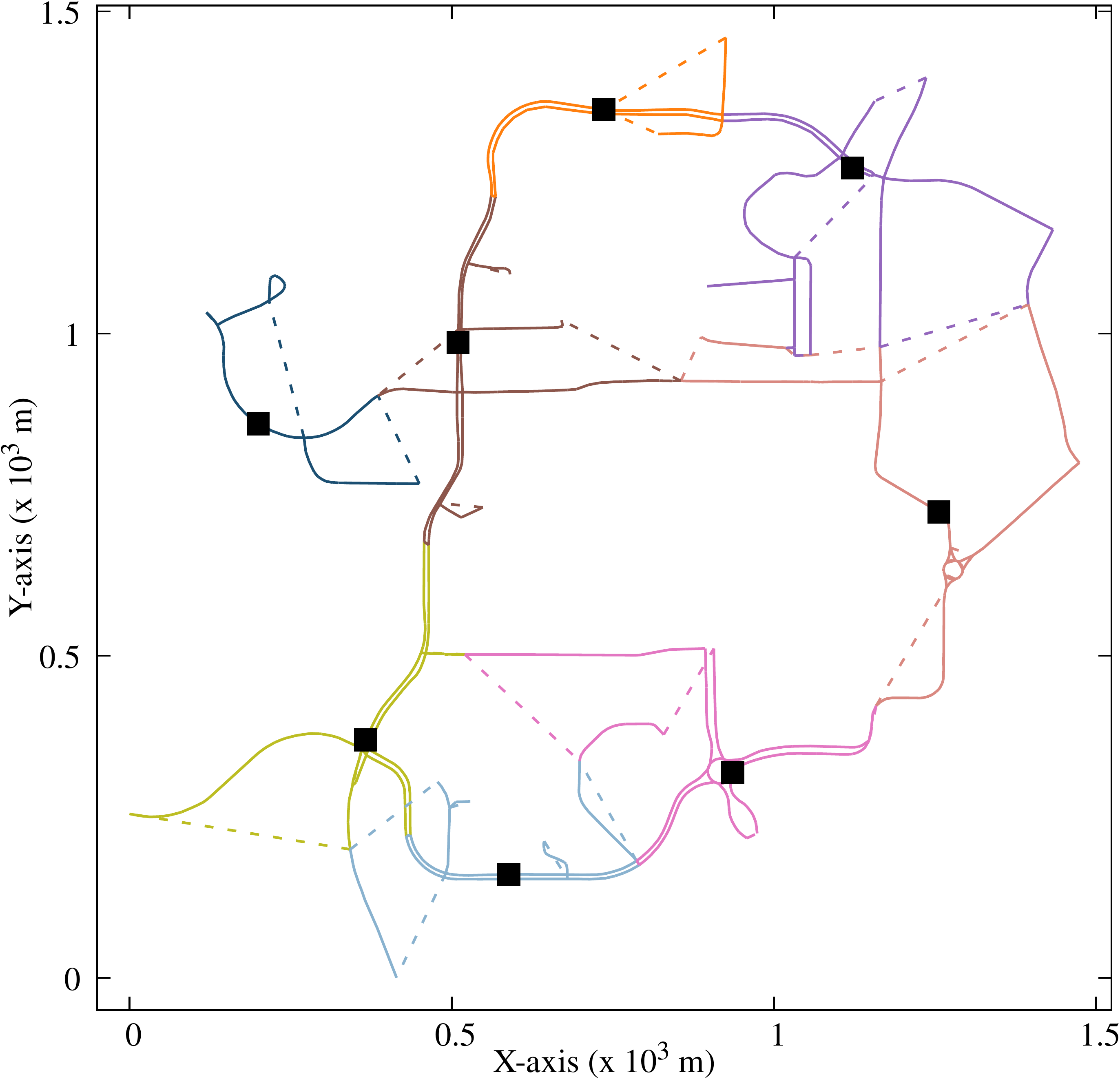}}
	\hfill
	\subfloat[Orthomosaic]{%
	\includegraphics[height=0.23\textheight]{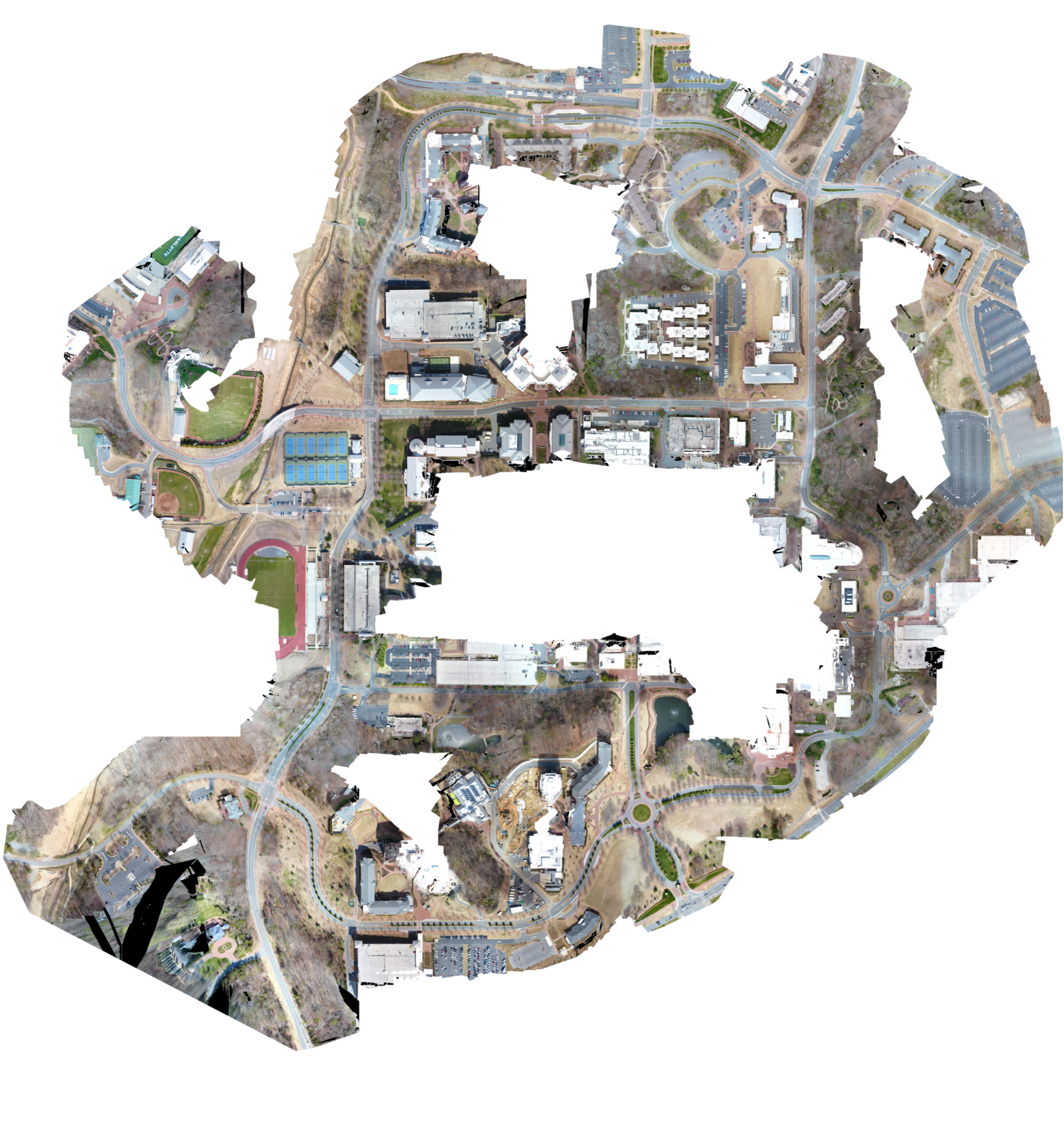}}
	\caption[Line coverage of the UNC Charlotte road network using a team of resource-constrained UAVs]{Line coverage of the UNC Charlotte road network using a team of resource-constrained UAVs.
		(a)~The input road network is 13\SIkm{} in length and spans an area of 1.5\SIkmsqr{}.
		The road network is modeled as a graph comprising 842 vertices and 865 required edges representing the road segments.
		As the UAVs can fly from one vertex to another, we add a non-required edge between each pair of vertices, resulting in 353,196 non-required edges in the graph.
		(b)~Eight routes for a team of UAVs are computed using the multi-depot Merge-Embed-Merge (MEM) algorithm developed in this paper.
		The algorithm computes depot locations, shown by black squares, from where the UAVs start and end their routes.
		The solid lines represent servicing, while the dashed lines represent deadheading.
		The UAVs can fly faster while deadheading, thereby optimizing total flight time.
		(c)~An orthomosaic of the road network generated from the images taken by the UAVs flown autonomously along the computed routes.
	\label{fig:uncc}}
\end{figure*}

\IEEEPARstart{L}{ine} coverage is the task of servicing linear environment features using sensors or tools mounted on robots.
The features to be serviced are modeled as one-dimensional segments (or curves), and all points along the segments must be visited.
Consider a natural disaster scenario such as flooding.
A team of uncrewed aerial vehicles (UAVs) is deployed to assess the accessibility of a road network for emergency services.
The UAVs must traverse the line segments corresponding to the road network and capture images for analysis.
This paper seeks to answer the following question: How should efficient routes for UAVs be planned so that each road network segment is serviced?
Mobile robots are often resource-constrained---they must come back to their {\em depot} or launch location before they exhaust their resources, such as battery life.
Figure~\ref{fig:uncc} shows a large road network, routes for eight UAVs covering the entire road network, and an orthomosaic generated from the images taken by the team of UAVs.
Power lines and gas pipelines have similar infrastructure that requires frequent inspection.
Additional applications arise in perimeter inspection and surveillance, traffic analysis of road networks, and welding and 3D printing operations.

Line coverage is closely related to arc routing problems (ARPs) studied in the operations research community~\cite{CorberanL14}.
ARPs have been used to generate routes for snow plowing, spraying salt, and cleaning road networks~\cite{CorberanEHPS21}.
The ARPs and their algorithms are designed specifically for human-operated vehicles.
Although the above tasks can potentially be automated with ground robots, line coverage has received limited attention in the robotics community.
In this paper, we design algorithms for line coverage using autonomous systems applicable to aerial, ground, and underwater robots.

The line coverage problem with multiple robots, modeled using a graph, has three defining attributes:
(1)~The edges in the graph are classified as required and non-required,
(2)~Robots have two modes of travel---servicing and deadheading, and
(3)~Robots have constraints on the resources available to them.
Required edges in the graph correspond to the line features to be covered, and a robot can use the non-required edges to travel from one vertex to another to reduce cost.
The vertices in the graph represent the endpoints of the edges.
A robot is said to be {\em servicing} a required edge when it performs task-specific actions such as collecting sensor data.
Each required edge needs to be serviced exactly once by any robot.
Robots may also traverse an edge without performing servicing tasks to optimize travel time, conserve energy, or reduce the amount of sensor data.
This mode of travel is known as {\em deadheading}, and both types of edges may be used any number of times for this purpose.

A service cost and a deadhead cost (e.g., travel time) are associated with each required edge, and they are incurred each time an edge is serviced or deadheaded, respectively.
Only the deadhead cost is associated with the non-required edges.
The total sum of the service and deadhead costs over all routes is to be minimized.
Moreover, traversing an edge results in the consumption of resources such as battery life.
These are modeled as {\em demands} on the edges, and the total demand of a route should be less than the given {\em resource capacity} of the robots.
The costs and demands for servicing an edge are usually more than those for deadheading, as task-related actions are only performed while servicing.
For example, a robot servicing an edge by recording images may travel slower to avoid motion blur, resulting in a longer travel time.

The {\em line coverage problem with multiple robots} is the task of computing efficient coverage routes for a set of line features such that the total cost of travel is minimized while respecting the resource constraints.
Starting with the line coverage problem for asymmetric graphs as the core problem, we formulate an integer linear program (ILP) and design a constructive heuristic algorithm called Merge-Embed-Merge (MEM).
To generate deployable solutions for the line coverage problem  in real-world scenarios, we address three important practical aspects:
(1)~Asymmetry in costs and demands,
(2)~Large graphs that require multiple depots, and
(3)~Turning costs and nonholonomic constraints.
This paper, for the first time, models these important practical aspects and presents a unified approach for the line coverage problem by developing highly efficient algorithms to compute routes for the robots.

{\bf Asymmetry in costs and demands:} In many robotics applications, the cost of travel and resource demands are direction-dependent.
For example, a ground robot traveling uphill can take longer and consume more energy than when traveling downhill.
Similarly, the costs and demands of aerial robots may differ in two directions due to wind conditions.
Hence, we consider the graph to have asymmetric cost and demand functions for servicing and deadheading.
Such asymmetric functions can also model one-way streets for ground robots.

{\bf Large graphs using multi-depot formulation:}  When the network is large, it may not be possible to service the entire network from a single {\em depot} (or home) location.
A standard preprocessing approach is to divide the input graph into small subgraphs using a clustering-based approach, assign a depot location for each cluster, and treat each subgraph as an independent single-depot problem.
However, these preprocessing steps require a {\em similarity} function for clustering, which does not typically consider the nuances of routing problems, such as the minimization of the route cost and the constraints on the resources, and so can lead to inefficient routes.
Furthermore, as the individual subgraphs are solved independently, routes that use edges from different subgraphs are not permitted, thereby significantly restricting the solution space.
To address large-scale graphs, this paper presents a multi-depot ILP formulation where the robots have the flexibility to optimize their routes by starting and ending their routes at one of several depots.
We further develop an efficient multi-depot MEM heuristic algorithm, MD-MEM, which optimizes the total cost of the routes, respects the resource constraints, and considers all the available depot locations for the routes.

{\bf Turning costs and nonholonomic constraints:}
Sharp turns can be costly for robots as they require the robots to slow down, turn, and then accelerate.
{\em Nonholonomic} robots, such as fixed-wing UAVs and underwater vehicles, cannot make sudden turns.
Often a postprocessing step is employed to modify computed routes so that nonholonomic constraints are respected.
Such procedures, however, are not guaranteed to comply with the resource capacity constraint.
To account for these, the paper directly incorporates turning costs and nonholonomic constraints into the multi-depot MEM algorithm.

While the practical aspects discussed above are sometimes addressed separately in motion planning and routing problems, they have not been addressed satisfactorily for the line coverage problem.
Often a graph preprocessing or route postprocessing step is employed to resolve the practical constraints.
To the best of our knowledge, this paper is the first to develop a unified approach to address the above three practical aspects of the line coverage problem.

The multi-robot line coverage problem is NP-hard, and there are no known efficient approximation algorithms with theoretical bounds, even for simpler variants of the problem.
Thus, we establish the efficacy and efficiency of the MEM heuristic algorithm in simulation on two new datasets\footnote{The dataset is available at:\\\url{https://github.com/UNCCharlotte-CS-Robotics/LineCoverage-dataset}} consisting of road networks from the 50 most populous cities in the world.
Illustrative physical experiments are also presented.
We provide an open-source implementation\footnote{The source code is available at:\\\url{https://github.com/UNCCharlotte-CS-Robotics/LineCoverage-library}} of our algorithms.

Building on our earlier publication~\cite{AgarwalA20ICRA}, this paper
(1)~simplifies and improves the ILP formulation,
(2)~addresses large graphs with multiple depots,
(3)~incorporates turning costs and nonholonomic constraints,
(4)~provides extensive simulation results, and
(5)~validates the algorithms in experiments for the multi-robot line coverage problem.
The conference paper addressed large environments by partitioning the input graph into smaller subgraphs and then solving for each subgraph.
However, the approach does not consider routing constraints in the partitioning step and, therefore, has several limitations that lead to inefficient routes due to the restricted solution space.
This paper addresses the multi-depot line coverage problem within the routing algorithm so that a preprocessing step for graph partitioning is no longer needed, thereby generating efficient routes.
We further incorporate turning costs and nonholonomic constraints into the formulation and the algorithms.
This leads to the first algorithm for the line coverage problem that has a unified approach for (1)~asymmetry in costs and demands, (2)~large graphs with multiple depots, and (3)~turning costs and nonholonomic constraints.

The rest of the paper is organized as follows.
The related work is discussed in \scref{sc:related_work}.
The multi-robot line coverage problem is formally described in \scref{sc:problem_statement}, along with ILP formulation.
\scref{sc:solution_approach} develops the constructive heuristic algorithm in stages, resulting in a unified and efficient approach.
The simulations and experiments are discussed in \scref{sc:sim}.
\scref{sc:conclusion} concludes the paper.

\section{Related Work}
\label{sc:related_work}

\begin{figure*}[htbp]
	\centering
	\hspace{0.7cm}
	\subfloat[Point Coverage]{%


\begin{tikzpicture}[scale=0.40]
	\node[fill=mBlue,rectangle] (0) at (4,4) {};
	\node[place,minimum size=1.7mm] (1) at (2,0) {};
	\node[place,minimum size=1.7mm] (2) at (8,0) {};
	\node[place,minimum size=1.7mm] (3) at (0,2) {};
	\node[place,minimum size=1.7mm] (4) at (1,1) {};
	\node[place,minimum size=1.7mm] (5) at (5,2) {};
	\node[place,minimum size=1.7mm] (6) at (8,2) {};
	\node[place,minimum size=1.7mm] (7) at (3,3) {};
	\node[place,minimum size=1.7mm] (8) at (6,3) {};
	\node[place,minimum size=1.7mm] (9) at (5,5) {};
	\node[place,minimum size=1.7mm] (10) at (8,5) {};
	\node[place,minimum size=1.7mm] (11) at (1,6) {};
	\node[place,minimum size=1.7mm] (12) at (2,6) {};
	\node[place,minimum size=1.7mm] (13) at (3,7) {};
	\node[place,minimum size=1.7mm] (14) at (6,7) {};
	\node[place,minimum size=1.7mm] (15) at (0,8) {};
	\node[place,minimum size=1.7mm] (16) at (7,8) {};
	\draw[thick, mGreen] (0) -- (7) -- (1) --(4) -- (3) -- (0);
	\draw[thick, mDarkRed] (0) -- (8) -- (6) --(2) -- (5) -- (0);
	\draw[thick, mOrange] (0) -- (12) -- (11) --(15) -- (13) -- (0);
	\draw[thick, mPurple] (0) -- (9) -- (14) --(16) -- (10) -- (0);
\end{tikzpicture}
	}
	\hfill
	\subfloat[Line Coverage]{%
		\includegraphics[width=0.25\textwidth,trim=1.4cm 1.0cm 0 0,clip]{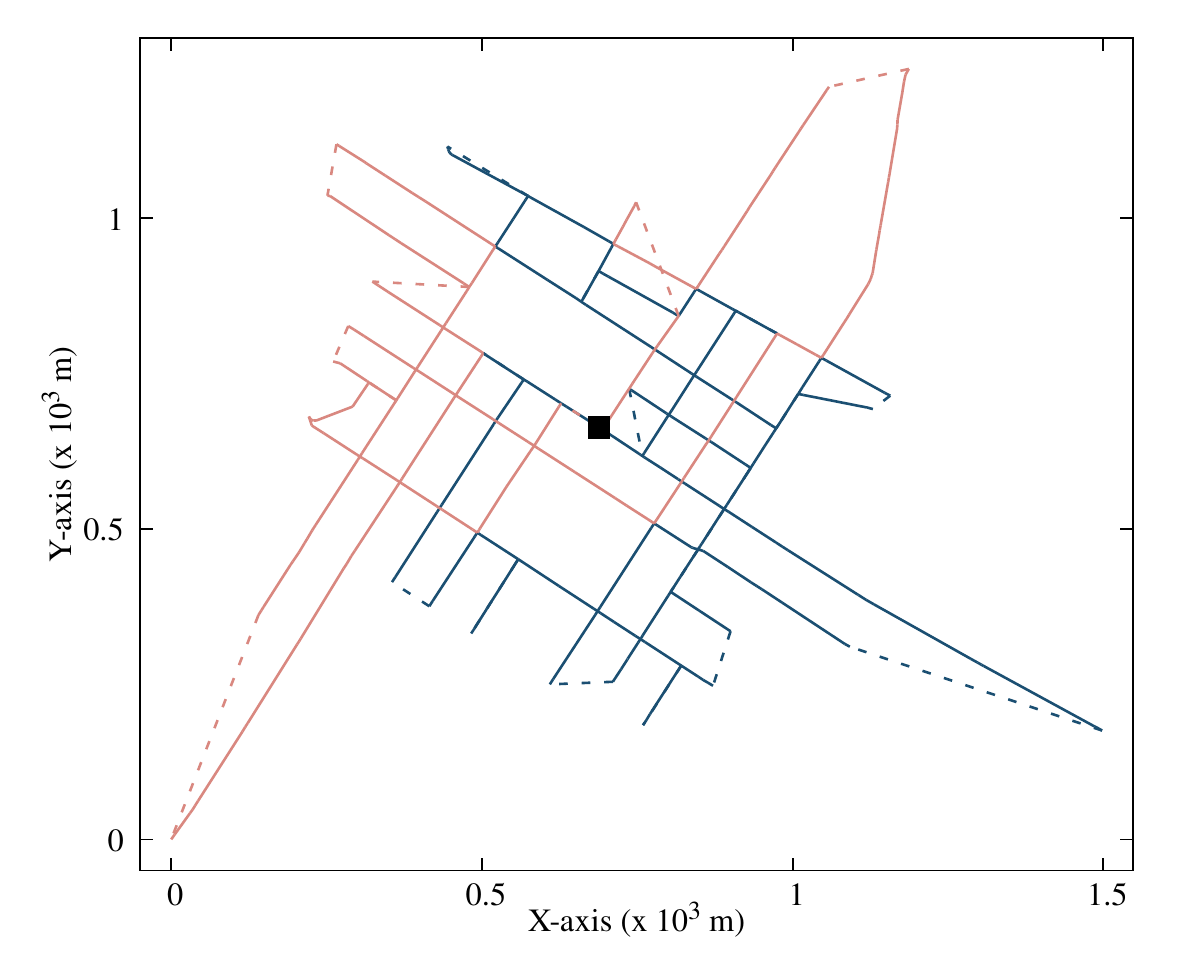}
	}
	\hfill
	\subfloat[Area Coverage]{%
		\includegraphics[width=0.3\textwidth]{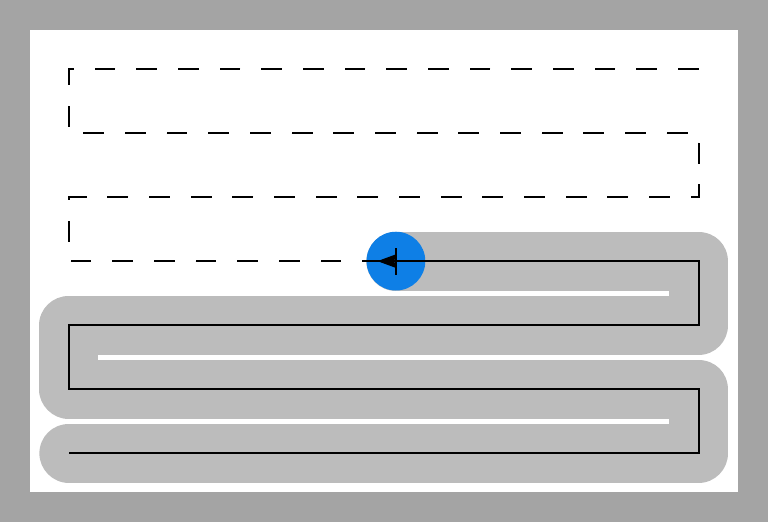}
	}
	\hspace{0.7cm}
	\caption[Types of features and the corresponding coverage problems]{%
		Three types of features and the corresponding coverage problems:
		(a)~Point coverage is the coverage of zero-dimensional point features and is commonly solved using node routing algorithms.
		(b)~Line coverage is the coverage of one-dimensional line features and belongs to the broad class of arc routing problems.
		(c)~Area coverage is the coverage of two-dimensional regions, often solved using computational geometry techniques.
	\label{fig:coverage_types}}
\end{figure*}
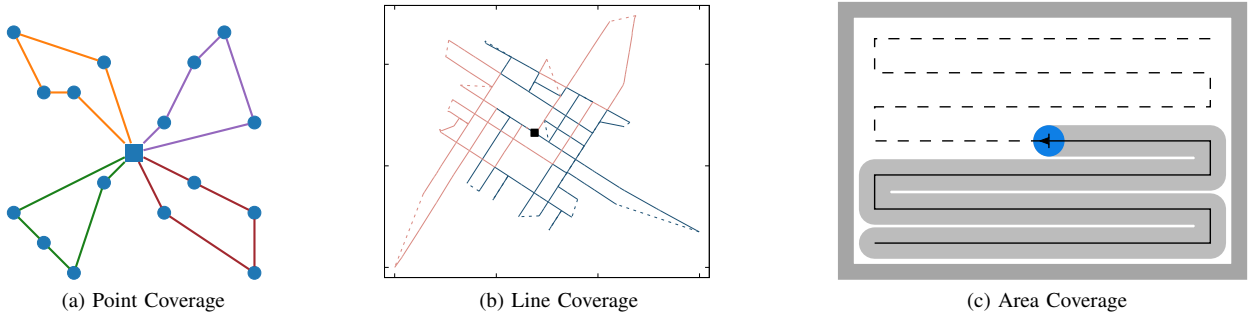

In a {\em coverage} application, the robots are required to visit specified features in the environment.
	Environments may have features of interest that can be represented as points, lines or curves, and areas, resulting in three distinct types of coverage problems, as illustrated in \fgref{fig:coverage_types}.
We provide a brief discussion on node routing problems, often used to solve the coverage of points.
Next, the arc routing problems (ARPs) and their relation to the line coverage problem are discussed.
Finally, the application of line coverage to area coverage problems in robotics is introduced.

\subsection{Point Coverage and Node Routing Problems}
Point coverage is the coverage of point features.
Single robot versions of the problem, without resource constraints, can be modeled on a graph and solved as a  traveling salesperson problem (TSP)~\cite{AlgorithmIlluminated4}.
Since deciding whether a graph is Hamiltonian is NP-complete~\cite{Karp72}, a reduction from the Hamiltonian cycle problem shows that the TSP is NP-hard, and approximating it to a constant factor is also NP-hard.
 Christofides~\cite{Christofides76} gave a 3/2-approximation algorithm for the TSP when the edge costs satisfy the triangle inequality.
When the edge costs are asymmetric, the problem is called the asymmetric traveling salesperson problem (ATSP); this is much harder than its symmetric counterpart.
Traub and Vygen~\cite{TraubV20ATSP} recently gave a $(22+\epsilon)$-approximation algorithm for the ATSP under triangle inequality conditions.
The line coverage problem can model points by treating each point as a degenerate edge, i.e., a required edge of zero cost, and connecting them to each other by non-required edges.
This makes the ATSP a special case of the line coverage problem, and the computational difficulty of the ATSP applies to the line coverage problem.

In the orienteering problem~\cite{GoldenLV87}, a reward is associated with the nodes, and the goal is to maximize the total reward while respecting a given resource constraint.
This problem is also NP-hard and approximation algorithms have been developed~\cite{VansteenwegenSO11,ChekuriKP12}.
Variations of the orienteering problem, such as with Dubins curves~\cite{XuLXPPLJD21}, multiple vehicles~\cite{PenickaFVS17,FaiglVP19}, and correlations on arcs~\cite{AgarwalA22WAFR} have also been studied.
However, the tour is not guaranteed to visit every node due to the resource constraint.
Hence, the orienteering problem cannot be used for point coverage, where {\em all} points must be visited.
In contrast, our formulation of the line coverage problem generates routes that together cover all the required edges and potentially required nodes (points) while ensuring each route respects the resource constraints.

There are additional variants of node routing problems, such as the generalized traveling salesperson problem, the vehicle routing problem~\cite{vrpbook}, the gas station problem~\cite{Khuller07}, and routing with fuel depots~\cite{Sundar14}.
However, they do not consider asymmetric costs, multiple depots, resource capacity, and turning costs within a unified framework.
Furthermore, they are fundamentally different from line coverage as they do not require the robot to traverse a set of required edges.

\subsection{Arc Routing Problems}
The line coverage problem belongs to the broad class of arc routing problems (ARPs).
The ARPs are usually applied to transportation problems in which servicing is related to tasks such as delivery and pick-up of goods~\cite{CorberanL14}.
In such problems, the costs are associated with the travel distances, and they have the same value for servicing and deadheading.
Furthermore, the demands are associated with the load the vehicles can carry, e.g., the capacity of a garbage collection truck.
Therefore, the deadheading of an edge does not affect the capacity.
In contrast, the capacity in the line coverage problem is associated with battery life, and demands are incurred in both modes of travel.

The Chinese postman problem (CPP), the simplest of the ARPs, is to find an optimal tour such that every edge in an undirected and connected graph is traversed at least once.
Edmonds and Johnson~\cite{EdmondsJ73} used matching and network flow to solve the CPP on undirected, directed, and Eulerian mixed graphs.
In the rural postman problem (RPP), the task is to service a subset of edges in a graph.
Frederickson~\cite{Frederickson79} gave a 3/2-approximation algorithm based on the algorithm given by Christofides~\cite{Christofides76} for the TSP.
The asymmetric RPP, i.e., RPP on a graph with asymmetric edge costs, is much harder than the symmetric counterpart.
The formulation for the single robot line coverage problem, given by Agarwal and Akella~\cite{AgarwalA20WAFR,AgarwalA23Networks}, can be used to solve asymmetric RPP.
They developed algorithms based on the structure of the required graph---the graph induced by the set of required edges.
A 2-approximation algorithm for the case of a connected required graph and a $(\alpha(C)+2)$-approximation algorithm for a required graph with $C$ connected components were developed.
Here, $\alpha(C)$ is the approximation factor for an algorithm for the ATSP with triangle inequality conditions on the edge cost.
The generalized routing problem (GRP) explicitly considers required nodes and required edges in the graph, and new heuristic algorithms have been proposed recently~\cite{BoyaciGRP23}.

Easton and Burdick~\cite{EastonB05} introduced $k$RPP, the RPP with $k$ vehicles.
They modeled coverage of 2D object boundaries as a $k$RPP and presented a cluster first and route-second heuristic.
The $k$RPP does not consider the capacity of the robot (vehicle).
The capacitated arc routing problem (CARP), introduced by Golden and Wong~\cite{GoldenW81}, considers $k$ vehicles of a given capacity $Q$.
The required edges have a demand associated with them, and the total demand of a route should be less than the capacity.
The CARP is a special case of the line coverage problem---there are no deadheading demands, and all costs and demands are symmetric.
Since the problem is NP-hard~\cite{GoldenW81}, several heuristic algorithms have been developed for the CARP~\cite{CorberanL14}.
W{\o}hlk~\cite{Wohlk08} presented a 7/2-approximation algorithm for the problem.
Several variants of the ARP partially or individually consider a subset of the characteristics of the line coverage problem.
Gouveia et al.~\cite{GouveiaMP10} presented a lower bound approach for the CARP on mixed graphs using a mixed integer linear program (MILP).
The CARP-DD, introduced by Kirlik and Sipahioglu~\cite{KirlikS12}, considers deadheading demands for the CARP.
The line coverage problem is a generalization of all these ARPs, as shown in Table~\ref{tb:arp_comparison}.
\begin{table}[htbp]
	\renewcommand{\arraystretch}{1.5}
	\caption{The line coverage problem with its special cases}
	\label{tb:arp_comparison}
	\centering
	\begin{centering}
		\begin{tabular}{@{}lcccc@{}}
			\toprule
			\multicolumn{1}{c}{\multirow{2}[3]{*}{Literature}} &  \multicolumn{2}{c}{Service}& \multicolumn{2}{c}{Deadheading}\\
			\cmidrule(lr){2-3} \cmidrule(lr){4-5}
																												 & Cost & Demand & Cost  & Demand \\ \midrule
			$k$RPP~\cite{EastonB05} & S & $-$ & $=$ & $-$ \\
			CARP~\cite{GoldenW81} & S & S & $=$ & $-$ \\
			MDCARP~\cite{MuyldermansCO03} & S & S & $=$ & $-$ \\
			CARP-DD~\cite{KirlikS12} & S & S & $=$ & S \\
			Mixed-CARP~\cite{GouveiaMP10} & A & A & A & $-$ \\
			{\bf Line coverage problem}& {\bf A} & {\bf A} & {\bf A} & {\bf A} \\
			\bottomrule
																 &&&&\\
		\end{tabular}
	\end{centering}

	S: symmetric, A: asymmetric, and $-$: not considered.\\
	=: The deadheading costs are equal to the service costs.
\end{table}

In the multi-depot capacitated arc routing problem (MDCARP), a set of depot locations is given as input.
The vehicles can start and end their route at any depot, and usually, the requirement is to return to the start depot at the end of the route.
The MDCARP is especially relevant in large-scale applications where servicing a network may not be optimal or even feasible from a single location.
Given the low battery life of robots, compared to fueled vehicles, the MDCARP becomes relevant even for moderately sized graphs.
One common technique is to cluster the input graph into smaller subgraphs and assign a single depot for each subgraph.
The algorithms for the CARP are then used on each of the subgraphs independently.
Such techniques have also been described as districting by Muyldermans et al.~\cite{MuyldermansCO03} and as sectoring by Mour{\~a}o et al.~\cite{MouraoNP09}.

Several exact and metaheuristic algorithms have been proposed for the ARPs.
They are covered in the survey paper by Corber{\'a}n and Prins~\cite{CorberanP10} and the monograph by Corber{\'a}n and Laporte~\cite{CorberanL14}.
Exact approaches include branch-and-bound with cutting planes, branch-and-price, and column generation.
The problem is NP-hard, and these exponential-time algorithms are not suitable for large-scale robotics applications.
Metaheuristic algorithms, such as scatter search, tabu search, and variable neighborhood descent, have been used for ARPs.
Similar to the exact methods, these can require significant computation resources.
Moreover, they typically require a good initial solution as an additional input to upper bound the optimal cost.
The heuristic algorithms presented in this paper can provide such an initial solution and be used as a fast sub-routine to generate intermediate solutions.

Algorithms based on minimum-cost perfect matching and minimum-cost flow problems are used widely to solve ARPs.
They generate a set of Eulerian digraphs, and a robot route can be generated by computing an Eulerian tour for each such digraph.
Incorporating multiple depots, turning costs, and nonholonomic constraints is a significant challenge when using these approaches.
In contrast, this paper develops a constructive heuristic algorithm, Merge-Embed-Merge (MEM), which maintains a set of routes for the robots and constructively merges pairs of routes to form larger routes.
The algorithm is very fast and gives solutions of high quality.
Using the basic structure of the algorithm for asymmetric costs, we solve the multi-depot version of the line coverage problem for large graphs and incorporate turning costs and nonholonomic constraints.
To the best of our knowledge, this is the first algorithm to address these challenging practical aspects of line coverage in a unified manner within a single algorithm.

\subsection{Line Coverage in Robotics}
There has been little work on using robots for line coverage tasks.
Dille and Singh~\cite{DilleS13} presented algorithms to perform coverage of a road network using a single aerial robot with kinematic constraints.
They modeled the problem through the tessellation of the road segments by circles of radius corresponding to the sensor footprint and computing a subset of the circular regions covering the entire road network.
Algorithms for node routing problems, such as the TSP and the multiple TSP, are then used to find the routes for the robots.
Oh et al.~\cite{OhKTW14} proposed an MILP formulation and a heuristic algorithm for the coverage of road networks.
The nearest insertion heuristic, originally designed for TSP, finds a sequence of edges to be visited while incorporating Dubins curves for nonholonomic robots.
The sequence is split across a team of robots using an auction algorithm.
Algorithms for  $k$RPP were developed for boundary inspection with multiple robots by Easton and Burdick~\cite{EastonB05}.
Williams and Burdick~\cite{WilliamsB06} developed algorithms for boundary inspection while considering revisions to the path plan for the robots to account for environmental changes.
Xu and Stentz~\cite{XuS10} use CPP and RPP formulations for environmental coverage and consider the case of incomplete prior map information.
They extended this work to multiple robots~\cite{XuS11ICRA} using $k$-means clustering to decompose the environment into smaller components, similar to the approach in~\cite{EastonB05}.
Campbell et al.~\cite{CampbellCPS18} presented an application of ARPs to cover road networks using a single UAV.
They discretize the required edges to allow the UAV to service an edge in parts.

The above work illustrates the line coverage applications in robotics.
However, line coverage has not been studied as extensively as the area coverage problem or node routing problems.
Current work does not usually consider the battery life of the robots and thus may not be suitable for large-scale applications.
In contrast, we model the battery life as capacity (resource) constraints and consider multiple depots, enabling solutions for large networks.
Furthermore, we allow demands on resources while deadheading and asymmetric functions for costs and demands.

\subsection{Arc Routing Problems in Area Coverage}
ARPs have been used in area coverage problems as a subroutine to generate efficient robot routes.
Arkin et al.~\cite{ArkinFM00} used the CPP to find a route for the milling problem, a variant of the area coverage problem wherein the tool is constrained within the workspace.
Mannadiar and Rekleitis~\cite{MannadiarR10} performed a cell decomposition and computed a Reeb graph.
The edges of the graph correspond to the cells in the decomposition.
The problem of visiting the cells was then formulated as the CPP.
Karapetyan et al.~\cite{KarapetyanBMTR17} used the CPP to compute a large Euler tour and then decomposed the tour into smaller ones using an algorithm given by Frederickson et al.~\cite{FredericksonHK76}.
Recently, we presented a new approach~\cite{AgarwalA22RAL} for the area coverage problem by generating service tracks in the environment after performing cell decomposition.
These service tracks were modeled as required edges for the line coverage problem, and the MEM algorithm was used to generate efficient routes for multiple resource-constrained robots.
Using the above technique, the algorithms presented in this paper for multiple depots and nonholonomic robots apply to the area coverage problem.

\section{The Line Coverage Problem}
\label{sc:problem_statement}
We pose the line coverage problem as an optimization problem on a graph.
The environment comprises linear features (line segments or curves) that need to be serviced by a homogeneous team of robots.
It is modeled as an undirected and connected graph $G=(V, E, E_r)$, where $E_r\subseteq E$ is the set of {\em required edges} representing the linear features.
The graph may have edges that do not require servicing, and the robots can use them to optimize their routes; these are the {\em non-required edges} given by $E_n=E\setminus E_r$.
The set $E$ can contain parallel edges, i.e., we allow for~$G$ to be a multigraph.
The set of vertices $V$ consists of edge endpoints, edge intersections, and depot locations.
The {\em depots} $V_d\subseteq V$ are a subset of vertices at which the robots start and end their routes.

For each edge $e$ in $E$ we associate two directional arcs $a_e$ and $\bar a_e$ that are opposite in direction to one another.
There are two modes of travel for a robot---servicing and deadheading.
A robot is said to be {\em servicing} an edge when it performs task-specific actions such as taking images along the edge.
We associate two binary variables $\svk{a_e}$ and $\svk{\bar a_e}$ with servicing an edge $e\in E_r$ by a route~$k$: if a robot, executing the route~$k$, services edge~$e$ in the direction $a_e$, then $\svk{a_e}$ is 1 and 0 otherwise; similarly for the direction $\bar a_e$.
Let $K$ be an upper bound on the total number of routes.
Since each required edge is required to be serviced exactly once, we have:
\begin{equation}
	\sum_{k=1}^K \Big(\svk{a_e} + \svk{\bar a_e}\Big) = 1,\quad \forall e\in E_r
\end{equation}
If a robot traverses an edge without servicing it, the robot is said to be {\em deadheading}, e.g., this occurs when a robot travels from its depot to an edge to be serviced.
We associate two non-negative integer variables $\ddk{a_e}$ and $\ddk{\bar a_e}$ with deadheading an edge $e\in E$ in route $k$.
The edges can be deadheaded any number of times.
The task is to compute a set of $K$ routes $\Uppi=\{\pi^1, \ldots,\pi^K\}$.
If we have $K$ robots, then each route is executed by a distinct robot.
If not, some robots may have to execute multiple routes to cover all the required edges.
The formulation does not restrict the number of robots.

Servicing an edge $e\in E_r$ in the direction $a_e$ incurs a service cost $\scost{a_e}$; similarly for the direction $\bar a_e$.
Analogously, deadheading an edge $e\in E$ incurs deadhead costs $\dcost{a_e}$ and $\dcost{\bar a_e}$.
A robot may need to travel slower while performing task-specific actions, resulting in higher costs for servicing than deadheading.
Thus, the two cost functions can differ.
The costs are associated with a minimization objective function of the optimization problem, such as total travel time.

The cost of a route $\pi^k \in \Uppi$ is given as:
\begin{equation}
	\begin{split}
		c(\pi^k) = &\sum_{e\in E_r} \Big[\scost{a_e}\,\svk{a_e} + \scost{\bar a_e}\,\svk{\bar a_e}\Big]\\
		+ &\sum_{e\in E\phantom{_r}}\Big[\dcost{a_e}\,\ddk{a_e} + \dcost{\bar a_e}\,\ddk{\bar a_e}\Big]\label{eqn:route_cost}
	\end{split}
\end{equation}

Each robot is constrained by a resource such as operation time, total travel distance, or battery life.
Such a constraint is represented by a {\em budget} or {\em capacity} $Q$ for each robot.
The consumption of resources is modeled by demand functions $\sdemand{a_e}$ for servicing and $\ddemand{a_e}$ for deadheading an edge $e\in E$ in the direction $a_e$.
The total demand incurred by a robot for a route $\pi^k$ must be less than the capacity:
\begin{equation}
	\begin{split}
		q(\pi^k) = &\sum_{e\in E_r} \Big[\sdemand{a_e}\,\svk{a_e} + \sdemand{\bar a_e}\,\svk{\bar a_e}\Big]\\
		+ &\sum_{e\in E\phantom{_r}}\Big[\ddemand{a_e}\,\ddk{a_e} + \ddemand{\bar a_e}\,\ddk{\bar a_e}\Big]\leq Q\label{eqn:capacity}
	\end{split}
\end{equation}

We consider the edge costs and demands for both servicing and deadheading to be direction-dependent, i.e., the graph is asymmetric.
For example, $\scost{a_e}$ can differ from $\scost{\bar a_e}$.
Asymmetry in graphs can occur when modeling wind for aerial robots or terrain for ground robots.
It also allows us to model one-way streets, directed graphs, and mixed graphs in general.
This can be achieved by setting the cost and demand in the prohibited direction to be a large constant.
The traversal of edges can be modeled using any function, such as constant velocity or cubic trajectories, and travel time can be used as the cost function.
Similarly, demands and capacity can be specified in terms of battery life.
Such functions can also incorporate wind and terrain information~\cite{MeiLHL06,FrancoB15}.
In general, the costs and demands are allowed to be arbitrary non-negative constants.
If we also have point features $V_f$ in the environment, we add an artificial edge $(v, v)$ for each point feature $v\in V_f$.
The cost and the resource demand for servicing such an artificial edge would be the same as that of servicing the point feature, and the cost and demand for deadheading is set to zero.
This transformation allows modeling both the point and the line features in the same formulation.

{\bf Definition:} Given an undirected and connected graph $G=(V, E, E_r)$, the {\em line coverage problem} is to find a set of coverage routes~$\Uppi$ that services each required edge in $E_r$ exactly once and minimizes the total cost of the routes, while respecting the resource constraints.

\subsection{Integer Linear Programming Formulation}
An integer linear program (ILP) is a formulation for optimization problems with integer variables, a linear objective function, and a set of linear constraints.
An ILP formulation provides a concise mathematical description of the problem, and solving the ILP gives an optimal solution to an instance of the problem if the instance has feasible solutions.

\subsubsection{Variables} We have the following variables for the ILP.
\begin{itemize}
	\item Binary service variables $\svk{a_e}, \svk{\bar a_e} \in \{0, 1\}$ for each required edge $e\in E_r$ and each route $k$.
	\item Integer deadheading variables $\ddk{a_e}, \ddk{\bar a_e} \in \mathbb N \cup \{0\}$ for each edge $e\in E$ and each route $k$.
	\item Integer flow variables $\zk{a_e}, \zk{\bar a_e}\in \mathbb N \cup \{0\}$ for each edge $e\in E$ and each route $k$.
		The flow variables are used in connectivity constraints to ensure that routes are connected to the depots.
\end{itemize}

For now, we assume that all the robots start and end their routes at the same depot location $v_0\in V$.
We will generalize the formulation to multiple depots in the following subsection.

\subsubsection{Routing Constraints}
The routing constraints ensure connectivity of a route to the depot and eliminate sub-tours.
We represent the assigned depot for each route $k$ by $v_d^k\in V_d\subseteq V$. For the single depot problem, considered in this subsection, $v_d^k = v_0$ for all $k$.

For ease of notation, we define the following sets:
\begin{align*}
	\mathcal A\,=\,\bigcup_{e\in E} &\{a_e, \bar a_e\},\quad
	\mathcal A_r\,=\,\bigcup_{e\in E_r} \{a_e, \bar a_e\},\\
	H(\mathcal A, v)\,=&\,\text{arcs in $\mc A$ with $v$ as the head, and}\\
	T(\mathcal A, v)\,=&\,\text{arcs in $\mc A$ with $v$ as the tail}
.\end{align*}
Here, $\mc A$ is the set of all arcs, and $\mc A_r$ is the set of arcs corresponding to required edges.

We have the following set of routing constraints for each route $k \in \{1,\ldots,K\}$:
\begin{align}
	&\sumS{a\in T(\mc A, v_d^k)} \zk{a} \quad=\quad\sumS{a\in \mc A_r}\svk{a}\label{eqn:flowDepot}\\
	&\sumS{a\in H(\mc A, v)}\zk{a} \;- \sumSr{a\in T(\mc A, v)} \zk{a} \;=\sumSr{a\in H(\mc A_r, v)}\svk{a}, \quad \forall v \in V\setminus \{v_d^k\}\label{eqn:edgeFlow}\\
	& \zk{a}\quad\leq\quad\sumS{a\in \mc A_r}\svk{a}, \quad \forall a\in \mc A\label{eqn:flowLimit2}\\
	& \zk{a}\quad\leq\quad\lvert E \rvert \ddk{a}, \quad \forall a\in \mc A\setminus \mc A_r\label{eqn:flowLimit3}\\
	& \zk{a}\quad\leq\quad\lvert E \rvert (\svk{a} +\ddk{a}), \quad \forall a\in \mc A_r\label{eqn:flowLimit1}\\
	&\sumS{a \in H(\mc A_r, v)}\svk{a}+\sumSr{a \in H(\mc A, v)}\ddk{a}\;\;=\;\;\sumS{a \in T(\mc A_r, v)}\svk{a} + \sumSr{a\in T(\mc A, v)}\ddk{a} ,\quad\forall v \in V\label{eqn:symmetry}
\end{align}
The constraints~\eqref{eqn:flowDepot} to \eqref{eqn:flowLimit1} are generalized flow constraints that together ensure the connectivity of the route to the depot and prohibit any sub-tours.
The variables $\zk{a}$ are flow variables for each edge direction.
Constraint~\eqref{eqn:flowDepot} defines the amount of flow being released from the depot vertex $v_d^k=v_0$, which acts as a source of the flow.
For any vertex $v$ (other than the depot vertex), a flow equal to the number of servicing arcs, with $v$ as the head, is absorbed by the vertex.
This is expressed in constraints~\eqref{eqn:edgeFlow}.
The amount of flow through an arc is limited by constraints~\eqref{eqn:flowLimit2} to \eqref{eqn:flowLimit1}.
An edge has a positive flow if and only if it is traversed.
Finally, the vertex symmetry constraints~\eqref{eqn:symmetry} ensure that the number of arcs entering a vertex is the same as the number of arcs leaving it.

\noindent{\bf ILP Formulation:}
The objective function of the line coverage problem is the total cost of the routes, with $K$ as an upper bound on the total number of routes.
We can now pose the line coverage problem as an optimization problem formulated as an ILP:

\noindent Minimize:
\begin{align}
	&\sum_{k=1}^{K} c(\pi^k) = \sum_{k = 1}^K\Big[\sum_{a\in \mc A_r}\scost{a}\svk{a} + \sum_{a\in \mc A}\dcost{a}\ddk{a}\Big]
\end{align}
\noindent subject to:
\begin{align}
	&\sum_{k=1}^K \Big(\svk{a_e} + \svk{\bar a_e}\Big) = 1,\quad \forall e\in E_r\\
	&q(\pi^k) = \sum_{a\in \mc A_r} \sdemand{a}\,\svk{a} + \sum_{a\in \mc A}\ddemand{a}\,\ddk{a} \leq Q,\quad \forall k\\
	&\text{Routing constraints \eqref{eqn:flowDepot} to \eqref{eqn:symmetry} for each robot }\, k\\
	&\svk{a} \in \{0, 1\},\quad \forall a\in \mc A_r\,\quad \forall k\\
	&\ddk{a}, \zk{a} \in \mathbb N \cup \{0\}, \quad \forall a\in \mc A,\quad \forall k
\end{align}

\subsection{Multi-Depot Formulation}
\newcommand{\xkd}{x^{k}_{d}}
In graphs spanning over a large area, it may be inefficient or even infeasible to service all the locations from a single location.
Hence, we develop a multi-depot formulation, wherein we are given a set of potential depot locations $V_d\subseteq V$, and a route can be assigned to any of the depots to start and end the route.
The problem can be formulated by adding assignment constraints---each route needs to be assigned exactly one depot location from $V_d$.

In general, the entire vertex set $V$ could be the set of potential depot locations.
However, this would increase the complexity of the problem significantly.
Instead, a smaller subset of vertices, ideally of size $K$ or less, can help in reducing the complexity while also providing high-quality solutions.
These locations can be selected from the field of operation based on terrain data or by clustering the vertices or the edges.

We introduce the binary variable $\xkd$, which is 1 when the depot $v_d\in V_d$ is assigned to route $k$, and 0 otherwise.
\begin{align}
	\sum_{v_d\in V_d} \xkd = 1, \quad \forall k\label{eqn:depot1}\\
	v_d^k = \sum_{v_d\in V_d} v_d \, \xkd, \quad \forall k\label{eqn:depotassign}\\
	\xkd \in \{0, 1\},\quad \forall v_d\in V_d,\, \forall k\label{eqn:trivial}
\end{align}
Constraints~\eqref{eqn:depot1} ensure that exactly one depot is assigned to each route.
Contraints~\eqref{eqn:depotassign} assign the depots to the routes---$v_d^k$ is set to the depot $v_d$ for which $\xkd$ is~1.

Routing constraints \eqref{eqn:flowDepot} and \eqref{eqn:edgeFlow} depend on the assigned depot~$v_d^k$.
The constraint~\eqref{eqn:flowDepot} is active only for the assigned depot, whereas constraints~\eqref{eqn:edgeFlow} are active for all the vertices except the assigned depot.
These can be resolved by multiplying the variable $\xkd$ to both sides of the constraint~\eqref{eqn:flowDepot}, and the expression $(1-\xkd)$ to both sides of constraints~\eqref{eqn:edgeFlow}.

However, this will result in a quadratic set of constraints, and the problem will become non-linear.
Although such quadratic constraints can be converted to linear constraints by introducing additional binary variables and large constants, it would significantly increase the size of the problem.
As the motivation for the multi-depot problem is to solve large instances, it is not beneficial to formulate the problem as an ILP, which becomes harder to solve for instances with a large number of variables and constraints.
The difficulty of solving such variants of the problem further motivates the development of versatile constructive heuristic algorithms that can efficiently solve the problem for large instances.

\section{Costructive Algorithms for Line Coverage}
\label{sc:solution_approach}
\newcommand{\mergeSym}{\uplus}
The line coverage problem and its variants are NP-hard problems.
Computing optimal solutions, e.g., using an ILP formulation, is usually feasible only for small instances.
This motivates us to develop heuristic algorithms to compute high-quality solutions for large instances.
Moreover, the algorithm is constructive---it maintains a set of feasible routes and iteratively merges a pair of routes to form a new larger route.
The overall algorithm for the line coverage problem with multiple robots is built in stages.
We first present in detail the Merge-Embed-Merge (MEM) algorithm for graphs with asymmetric costs and demands.
Using the constructive nature of the MEM algorithm, we then develop the multi-depot MEM (MD-MEM) algorithm for large graphs.
Finally, we incorporate turning costs and nonholonomic constraints to develop a unified MD-MEM-Turns algorithm.

\subsection{Merge-Embed-Merge: A Constructive Heuristic}
This section develops a new algorithm, Merge-Embed-Merge (MEM), for the line coverage problem.
The underlying concept is to maintain a set of feasible routes; initially, a route is created for each required edge.
Subsequently, routes are merged together greedily to form a smaller set of routes.
This concept of merging was first presented by Clarke and Wright~\cite{ClarkeW64} for the capacitated vehicle routing problem, which was later adopted in the {\em Augment-Merge} heuristic by Golden et al.~\cite{GoldenDB83} for the CARP.
However, the Augment-Merge algorithm cannot handle the asymmetric costs and demands and the deadheading demands of the line coverage problem.
Furthermore, the heuristic degrades rapidly with instance size, especially when the set of required edges is much smaller than the entire edge set~\cite{CorberanL14}.
We improve this by including an {\em embed} step in our algorithm.

The MEM algorithm, given in \algref{alg:mem}, comprises four components:
(1)~{\em initialization} of routes,
(2)~computation of {\em savings},
(3)~{\em merging} two routes to form a new route, and
(4)~{\em embedding} the newly merged route.
A max-heap data structure is used to greedily decide the two routes to be merged and embed new routes.

\begin{figure}[htpb]
	\centering
\begin{center}
	\begin{tikzpicture}[scale=1]
		\small
		\usetikzlibrary{calc}
		\tikzstyle{scEnd}=[near end, fill=none, mDarkRed]
		\tikzstyle{scSt}=[near start, fill=none, black]
		\tikzstyle{scMid}=[midway, fill=none, black]


		\coordinate (0) at (0,0);
		\coordinate (da) at (-0.1,-0.1);
		\coordinate (db) at (0.1,0.1);
		\def\radi{3}

		\coordinate (1) at (210:\radi)		{};
		\coordinate (2) at (180:\radi)		{};
		\coordinate (3) at (150:\radi)		{};
		\draw let \p{A}=(3), \p{B}=(110:\radi) in coordinate (4) at (\x{B}, \y{A});
		\draw let \p{A}=(4), \p{B}=(70:\radi) in coordinate (5) at (\x{B}, \y{A});
		\coordinate (6) at (30:\radi)		{};
		\coordinate (7) at (0:\radi)		{};
		\coordinate (8) at (-30:\radi)		{};
		\draw let \p{A}=(8), \p{B}=(-90:\radi) in coordinate (dd) at (\x{B}, \y{A});

		\node[fill=none]	(0)			at (0)	{$R_p$};
		\node[fill=none, below=0.1]	(d)			at (dd)	{$v_0$};
		\node[fill=none, below left]		at (1)		{$i$};
		\node[fill=none, below right]		at (8)		{$j$};

		\draw[mid2arc=0.7, mBlue] (1) arc (210:180:\radi) node [scMid, below left] {$a_s$};
		\draw[mid2arc=0.7, mBlue] (3) -- (4) node [scMid, above] {$a_t$};
		\draw[mid2arc=0.7, mBlue] (5) -- (6) node [scMid, above] {$a_{t+1}$};
		\draw[mid2arc=0.7, mBlue] (7) arc (0:-30:\radi) node [scMid, below right] {$a_l$};
		\draw[mid2arc=0.7, mGreen, dashed] (dd) -- (1);
		\draw[mid2arc=0.7, mGreen, dashed] (2) arc (180:150:\radi);
		\draw[mid2arc=0.7, mGreen, dashed] (4) -- (5);
		\draw[mid2arc=0.7, mGreen, dashed] (6) arc (30:0:\radi);
		\draw[mid2arc=0.7, mGreen, dashed] (8) -- (dd);

		\fill[black] ($ (dd) + (da) $) rectangle ($ (dd) + (db) $);

		\foreach \x in {1,...,8} {
			\node[place] (\x) at (\x)			{};
		}
	\end{tikzpicture}
\end{center}
	\caption[Representation of a route]{Representation of a route $R_p$ as a sequence of arcs corresponding to required edges:
	The required arcs are shown as solid blue lines, and deadheadings are shown as dashed green lines.
	The route internally includes deadheadings given by shortest paths to and from the depot $v_0$ (black square).
The route may have deadheadings between two non-adjacent required arcs.}
	\label{fig:route}
\end{figure}
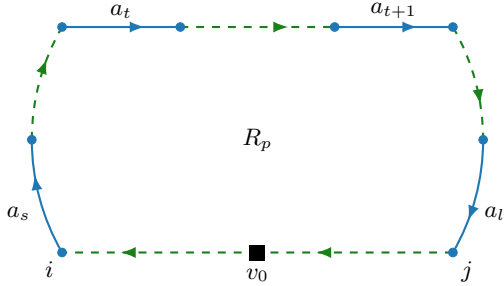

We begin by discussing the {\em representation of routes} in the algorithm.
A route is represented by a sequence of required arcs that are to be serviced by the robot traversing the route.
As the costs and the demands can be direction-dependent, arcs are used instead of the corresponding required edges.
The robot starts at the depot, travels to the starting vertex of the first required arc, services the sequence of required arcs, and returns to the depot.
Consider a route $R_p$, as shown in \fgref{fig:route}: the vertex $i$ is the starting vertex of the first required arc $a_s$, and the vertex $j$ is the end vertex of the last required arc $a_l$ in the sequence given by $R_p$.
The robot will deadhead from the depot $v_0$ to vertex $i$, service the required arcs starting from $i$ up to the last required arc ending at vertex~$j$, and then deadhead back to the depot $v_0$.
Note that the paths $v_0\rightarrow i$ and $j\rightarrow v_0$ are the shortest paths and may contain several arcs, all deadheaded.
If two successive required arcs $a_t$ and $a_{t+1}$ are not adjacent, the robot deadheads along the shortest path from the end vertex of $a_t$ to the start vertex of $a_{t+1}$.
Thus, the sequence $i \rightarrow j$ may involve deadheading between the constituent required arcs.
The cost of the route $R_p$ is:
\begin{align}
	c(R_p) = \dcost{v_0, i} + \scost{R_p} + \dcost{j, v_0} + \lambda
	\label{eq:}
\end{align}
The shortest path gives the costs of deadheading from and to the depot.
The cost of servicing the sequence of required arcs in $R_p$ is given by  $\scost{R_p}$, which may include deadheadings between non-adjacent required arcs.
An additional constant cost $\lambda$ representing route setup cost is added to the route cost.
The route setup cost represents the additional time required to launch a robot from a depot and helps in reducing the number of routes during the merging process.

\begin{algorithm}[htbp]
	\small
	\LinesNumbered
	\SetKwInOut{Input}{Input}\SetKwInOut{Output}{Output}
	\SetKwRepeat{Do}{do}{while}%
	\SetKw{Continue}{continue}
	\SetKw{Break}{break}
	\SetKw{KwAnd}{and}
	\SetKw{KwOr}{or}
	\Input{$G=(V, E, E_r)$, costs, demands, capacity $Q$}
	\Output{Coverage routes $\mc R$, where each tour $R\in \mc R$ is a sequence of required edges}
	\tcc{Initialization of routes}
	$\mc R\gets\emptyset$; $k \gets 1$\;
	\For(\tcp*[f]{$a_e$ and $\bar a_e$ are arcs for $e$}){$e \in E_r$\label{alg:mem:init_start}}{
		$i\gets \operatorname{tail}(a_e)$;
		$j\gets \operatorname{head}(a_e)$\;\label{alg:mem:tailhead}
		$c \gets  \dcost{v_0, i} + \scost{a_e} + \dcost{j, v_0} + \lambda $\;
		$\bar c \gets \dcost{v_0, j} + \scost{\bar a_e} + \dcost{i, v_0} + \lambda $\;
		\leIf{$c\leq \bar c$}{$R_k\gets a_e$}{$R_k\gets {\bar a_e}$\label{alg:mem:init_if}}
		$\mc R.\operatorname{push}(R_k)$; $k\gets k+1$\;\label{alg:mem:init_end}
	}
	\tcc{Compute savings}
	$S\gets\emptyset$\;
	\ForEach{pair of tours $R_p, R_q$\label{alg:mem:pairsavings}}{
		Compute saving $s_{pq}$ for $R_p\mergeSym R_q$\;
		\uIf{$s_{pq}\geq 0$ \KwAnd $ \operatorname{demand}(R_p\mergeSym R_q)\leq Q$}{
		$S.\operatorname{push}\big((p, q, s_{pq})\big)$\;\label{alg:mem:savingsend}}
	}
	$\operatorname{make\_heap}(S)$\tcp*{max-heap}\label{alg:mem:heap}
	\tcc{Repeated Merge and Embed}
	\While{$S\neq \emptyset$\label{alg:mem:whilestart}}{
		$(p, q, s)\gets S.\operatorname{extract-max}(\;)$\;\label{alg:mem:extract}
		\uIf{$R_p \neq \emptyset~\KwAnd~R_q \neq\emptyset$\label{alg:mem:mergecheck}}
		{ \tcc{Merge}
			$R_k\gets R_p\mergeSym R_q$\;\label{alg:mem:merge}
			$\mc R.\operatorname{push}(R_k)$; $k\gets k+1$\;\label{alg:mem:Rpush}
			$R_p\gets \emptyset$; $R_q\gets \emptyset$\;\label{alg:mem:setempty}
			\tcc{Embed}
			\ForEach{tour $R_i$ with $i\neq k~\KwAnd~R_i \neq \emptyset$\label{alg:mem:embedstart}}{
				Compute saving $s_{ki}$ for $R_k\mergeSym R_i$\;
				\uIf{$s_{ki}\geq 0$ \KwAnd $\operatorname{demand}(R_k\mergeSym R_i)\leq Q$\label{alg:mem:embedif}}{
					$S.\operatorname{insert}\big((k, i, s_{ki})\big)$\label{alg:mem:whileend}\;\label{alg:mem:embedend}
				}
			}
		}
	}
	Remove empty routes from $\mc R$\;
	\caption{Merge-Embed-Merge (MEM)}
	\label{alg:mem}
\end{algorithm}

\subsubsection{Initialization of Routes}
The MEM algorithm (Algorithm~\ref{alg:mem}) constructs a route for each required edge in the initialization step (lines  \ref{alg:mem:init_start}--\ref{alg:mem:init_end}).
Each edge $e\in E_r$ has two arcs $a_e$ and $\bar a_e$ associated with it, representing the two travel directions.
Let $i$ and $j$ denote the tail and the head of the arc $a_e$, respectively (line~\ref{alg:mem:tailhead}).
Then the route for servicing $a_e$ comprises the shortest path from the depot $v_0$ to the tail vertex~$i$, servicing of arc $a_e$, and the shortest path from the head vertex $j$ to the depot.
In the other direction, the route comprises the shortest path from the depot to vertex~$j$, servicing of arc $\bar a_e$, and the shortest path from vertex~$i$ to the depot.
Since the costs are asymmetric, of the two routes, the one with the lower cost is selected (line~\ref{alg:mem:init_if}).
It is assumed that the demand for the initial routes is less than the capacity; otherwise, the instance does not have a feasible solution.
A constant route setup cost~$\lambda$ is added to the route cost.
The routes are stored in the list~$\mc R$, and there are $m=\lvert E_r\rvert$ routes initially.
The particular case where an edge can be serviced in only one direction can be handled appropriately in the initialization step.

\subsubsection{Computation of Savings}
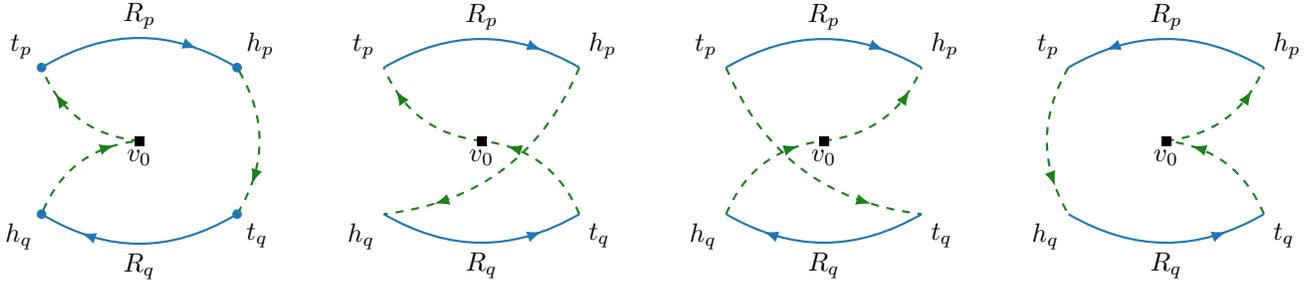
\begin{figure*}[htbp]
	\centering
	\begin{center}
	\begin{tikzpicture}[scale=0.65]
		\usetikzlibrary{calc}
		\tikzstyle{scEnd}=[near end, fill=none, mDarkRed]
		\tikzstyle{scSt}=[near start, fill=none, black]
		\tikzstyle{scMid}=[midway, fill=none, black]


		\coordinate (0) at (0,0);
		\coordinate (da) at (-0.1,-0.1);
		\coordinate (db) at (0.1,0.1);

		\node[place] (1) at (-2.0,  1.5)		{};
		\node[place] (2) at ( 2.0,  1.5)			{};
		\node[place] (3) at (-2.0, -1.5)		{};
		\node[place] (4) at ( 2.0, -1.5)		{};

		\node[fill=none, below]	(d)			at (0,0)	{$v_0$};
		\node[fill=none, above left]		at (1)		{$t_{p}$};
		\node[fill=none, above right]		at (2)		{$h_{p}$};
		\node[fill=none, below left]		at (3)		{$h_{q}$};
		\node[fill=none, below right]		at (4)		{$t_{q}$};

		\path[mBlue] (1)  edge [mid2arc, bend left]  node[scMid, above]  {$R_p$} (2);
		\path[mBlue] (4)  edge [mid2arc, bend left]  node[scMid, below]  {$R_q$} (3);
		\draw[mid2arc, mGreen, dashed] (0) to [bend left] (1);
		\draw[mid2arc, mGreen, dashed] (2) to [bend left] (4);
		\draw[mid2arc, mGreen, dashed] (3) to [bend left] (0);

		\fill[black] ($ (0) + (da) $) rectangle ($ (0) + (db) $);

		\coordinate (off) at (7, 0);
		\foreach \x in {1,...,4} {
			\coordinate (\x) at ( $ (\x) +  (off) $);
		}
		\node[fill=none, above left]		at (1)		{$t_{p}$};
		\node[fill=none, above right]		at (2)		{$h_{p}$};
		\node[fill=none, below left]		at (3)		{$h_{q}$};
		\node[fill=none, below right]		at (4)		{$t_{q}$};

		\coordinate (0) at ($ (0) + (off) $);
		\node[fill=none, below]	(d)			at (0)	{$v_0$};

		\draw[mid2arc, mGreen, dashed] (off) to [bend left] (1);
		\draw[mid2arc, mGreen, dashed] (2) to [bend left] (3);
		\draw[mid2arc, mGreen, dashed] (4) to [bend right] (off);
		\fill[black] ($ (off) + (da) $) rectangle ($ (off) + (db) $);
		\path[mBlue] (1)  edge [mid2arc, bend left]  node[scMid, above]  {$R_p$} (2);
		\path[mBlue] (3)  edge [mid2arc, bend right]  node[scMid, below] {$R_q$} (4);
	
		\foreach \x in {1,...,4} {
			\coordinate (\x) at ( $ (\x) +  (off) $);
		}
		\node[fill=none, above left]		at (1)		{$t_{p}$};
		\node[fill=none, above right]		at (2)		{$h_{p}$};
		\node[fill=none, below left]		at (3)		{$h_{q}$};
		\node[fill=none, below right]		at (4)		{$t_{q}$};

		\coordinate (0) at ($ (0) + (off) $);
		\node[fill=none, below]	(d)			at (0)	{$v_0$};
		\draw[mid2arc, mGreen, dashed] (0) to [bend right] (2);
		\draw[mid2arc, mGreen, dashed] (1) to [bend right] (4);
		\draw[mid2arc, mGreen, dashed] (3) to [bend left] (0);
		\fill[black] ($ (0) + (da) $) rectangle ($ (0) + (db) $);
		\path[mBlue] (1)  edge [mid2arc, bend left]  node[scMid, above]  {$R_p$} (2);
		\path[mBlue] (4)  edge [mid2arc, bend left]  node[scMid, below]  {$R_q$} (3);

		\foreach \x in {1,...,4} {
			\coordinate (\x) at ( $ (\x) +  (off) $);
		}
	
		\node[fill=none, above left]		at (1)		{$t_{p}$};
		\node[fill=none, above right]		at (2)		{$h_{p}$};
		\node[fill=none, below left]		at (3)		{$h_{q}$};
		\node[fill=none, below right]		at (4)		{$t_{q}$};

		\coordinate (0) at ($ (0) + (off) $);
		\node[fill=none, below]	(d)			at (0)	{$v_0$};
		\draw[mid2arc, mGreen, dashed] (0) to [bend right] (2);
		\draw[mid2arc, mGreen, dashed] (1) to [bend right] (3);
		\draw[mid2arc, mGreen, dashed] (4) to [bend right] (0);
		\fill[black] ($ (0) + (da) $) rectangle ($ (0) + (db) $);
		\path[mBlue] (2)  edge [mid2arc, bend right]  node[scMid, above]  {$R_p$} (1);
		\path[mBlue] (3)  edge [mid2arc, bend right]  node[scMid, below]  {$R_q$} (4);
	\end{tikzpicture}
\end{center}
	\caption[Four of eight permutations to merge two routes]{%
The figure shows four of the eight permutations to merge two routes $R_p$ and $R_q$.
The remaining four permutations consist of the shown permutations in the reverse directions.
The tail and the head vertices for $R_p$ are $t_p$ and $h_p$, respectively.
Similarly, $t_q$ and $h_q$ are defined for $R_q$.
The first merged route is $v_0\rightarrow t_p \xrightarrow{R_p} h_p\rightarrow t_q\xrightarrow{R_q} h_q\rightarrow v_0$ and its reverse direction route is $v_0\rightarrow h_q\xrightarrow{\overline R_q} t_q\rightarrow h_p\xrightarrow{\overline R_p} t_p\rightarrow v_0$.
The savings for merging two routes come from potentially reduced deadheading to and from the depot.}
	\label{fig:eightways}
\end{figure*}

Consider two routes $R_p$ and $R_q$, with tail and head vertices given by $t_p, h_p$ and $t_q, h_q$, as potential candidates for merging.
There are eight possible permutations to merge the two routes, of which four are shown in \fgref{fig:eightways}.
The remaining four ways consist of routes in the reverse directions.
The first merged route $R_{pq}$ and its cost $\operatorname{cost}(R_{pq})$ are:
\begin{align*}
	R_{pq}:= &R_p \mergeSym R_q:=v_0\rightarrow t_p \xrightarrow{R_p} h_p\rightarrow t_q\xrightarrow{R_q} h_q\rightarrow v_0\\
	\operatorname{cost}(R_{pq})=\,&\dcost{v_0, t_p} + \scost{R_p} + \dcost{h_p, t_q}\\
	+\,&\scost{R_q} + \dcost{h_q, v_0} + \lambda
	\label{eq:savings}
\end{align*}

Such a merge can have potential {\em saving} in cost since we no longer require $h_p\rightarrow v_0$ and $v_0\rightarrow t_q$.
There is also a cost-saving of a route setup cost~$\lambda$ as we have a single route instead of two.
Thus, the net saving in cost $s_{pq}$ for merging routes $R_p$ and $R_q$ is given by $\operatorname{cost}(R_p) + \operatorname{cost}(R_q) - \operatorname{cost}(R_{pq})$.
However, the cost savings are affected by the direction of the edges due to asymmetry in the costs.
Hence, we need to consider all eight permutations for merging two routes.
Some of these permutations might not satisfy the resource capacity constraint, i.e., the total demand of the merged route may be more than the capacity~$Q$.
The merged route that satisfies the resource capacity constraint and has the maximum cost savings is denoted by $R_p\mergeSym R_q$.
If no such combination exists, then $R_p \mergeSym R_q =\emptyset$ and savings $s_{pq} =-\infty$.
Note that the algorithm allows reversal of the service directions, i.e., the service arcs of a constituent route $R_p$ can be reversed in the merged route $R_{pq}$.
Hence, the direction of the arcs in the initialization step is not fixed, and the algorithm chooses the direction that yields the maximum cost savings in each iteration.

For each pair of routes that yield a feasible merged route, the maximum saving in cost is computed and stored as a tuple $(p, q, s_{pq})$, where $p$ and $q$ correspond to the routes considered, and $s_{pq}$ is the corresponding saving (lines~\ref{alg:mem:pairsavings}--\ref{alg:mem:savingsend}).
These $m(m-1)/2$ tuples are stored in a binary max-heap data structure~$S$, which can be built in $\mc O(m^2)$ computation time (line~\ref{alg:mem:heap}).

Next, the merge and embed steps are executed repeatedly until no further merges are possible (lines~\ref{alg:mem:whilestart}--\ref{alg:mem:whileend}).
\subsubsection{Merge}
The pair of routes with maximum cost savings is selected to form a new merged route, thereby maximizing immediate savings.
The maximum element from the max-heap~$S$ is extracted (line~\ref{alg:mem:extract}), and the constituent routes are merged if both are non-empty (lines~\ref{alg:mem:mergecheck}--\ref{alg:mem:merge}).
The merged route $R_k$ is added to the list of routes~$\mc R$ (line~\ref{alg:mem:Rpush}).
The constituent routes are set to $\emptyset$ so that they are no longer considered for future merges (line~\ref{alg:mem:setempty}).
Setting a route to $\emptyset$ does not modify the max-heap data structure, but only the corresponding element in the list of routes~$\mc R$.
Empty routes are filtered when the elements from the max-heap are extracted (line~\ref{alg:mem:mergecheck}).
The complexity of the merge step is $\mc O(\log\lvert S\rvert)$, where $\lvert S\rvert$ is the number of elements in~$S$.

\subsubsection{Embed}
We consider merging existing non-empty routes with the newly merged route $R_k$ in the embed step (lines~\ref{alg:mem:embedstart}--\ref{alg:mem:embedend}).
Potential cost savings are computed for the new route $R_k$ if merged with the other non-empty routes in the list $\mc R$.
New tuples $(k, i, s_{ki})$ are generated and inserted into the max-heap~$S$, if merging satisfies the capacity constraint and the cost saving is non-negative (lines~\ref{alg:mem:embedif}--\ref{alg:mem:embedend}).
As there are $\lvert\mc R\rvert -1 $ such new tuples, the embed step has a computational complexity of $\mc O\left(\left\lvert\mc R\right\rvert \log\left(\lvert S\rvert + \lvert \mc R\rvert\right)\right)$.

The merge and embed components are executed until no further merges are possible, i.e., $S=\emptyset$.
The maximum number of routes in the list $\mc R$ is upper bounded by $2m$, with at most $m$ non-empty routes at any iteration.
Here, $m=\lvert E_r\rvert$ is the number of required edges.
The maximum number of elements in the max-heap $S$ is $\mc O(m^2)$.
Thus the complexity of the repeated merge-embed component over all possible merges is $\mc O(m^2\log m)$, which is also the overall complexity of the algorithm.
Depending on the instance, one may need to compute the shortest deadheading paths between all pairs of vertices.
This can be done using the Floyd-Warshall algorithm in $\mc O(\lvert V\rvert^3)$ computation time~\cite{DasguptaPV06book}.
The MEM algorithm does not have an approximation guarantee; however, we show in \scref{sc:sim} that the algorithm performs well in practice.

There are two essential characteristics of the algorithm that have practical benefits:
(1)~The algorithm maintains a feasible set of routes.
These routes can be extracted at any point in the algorithm, giving it the {\em anytime} property.
(2)~The number of routes generated by the algorithm does not depend on the number of robots, i.e., the algorithm is agnostic to the number of robots available.
It generates the number of routes required to cover the entire environment completely.
Thus if a small fleet of robots is available, one can execute multiple routes for some of the robots by replacing or recharging batteries.

\subsection{Multi-Depot Line Coverage for Large Graphs: MD-MEM}
\label{sec:mdmem}
In the above MEM algorithm, we considered a single depot location where all the robots start and end their routes.
However, it may not be efficient or feasible to service the required edges representing the linear features from a single location for environments spanning over a large area.
Thus, we extend the MEM algorithm to enable multiple depots for the line coverage problem.


A common approach for solving multi-depot problems is to cluster the required edges and create subgraphs~\cite{MuyldermansCO03,MouraoNP09}; this is, in fact, what we did in our earlier work~\cite{AgarwalA20ICRA}.
Each subgraph is then treated as an instance of the single-depot problem.
However, as the generated clusters are treated independently, sometimes multiple routes are generated for a cluster with one of the routes serving only a small number of edges, i.e., the robots are not using their capacity to their full potential. The algorithm does not have the freedom to exchange or transfer some of these edges to nearby clusters. This often leads to inefficient solutions.
This limitation is resolved in the multi-depot version of the MEM algorithm by directly considering the depots in the routing process.

We are given as input a set of depot locations $V_d\subseteq V$.
These locations can be specified based on operator ease and field constraints, e.g., an operator may prefer to launch aerial robots from high vantage points.
Alternatively, the locations can be determined using a $k$-medoids clustering algorithm.
As clustering algorithms are usually non-deterministic, one may choose to perform clustering multiple times to obtain a more desirable set of depot locations.
Note that while the clustering may be used for assigning depot locations, it is not used to cluster the required edges.

Two modifications need to be made to the MEM algorithm given in Algorithm~\ref{alg:mem} to integrate the depot selection for the multi-depot line coverage problem:
(1)~The initialization of the routes, and
(2)~The computation of savings for merging two routes.

\subsubsection{Initialization of routes}
The initialization process for the multi-depot line coverage problem is given in the procedure~\ref{alg:init_md}.
For each required edge, we iterate over all the depot locations and compute the cost of servicing the edge in the two directions.
The number of such computations is $2\lvert V_d\rvert$.
The one with the lowest cost is selected, provided the demand is less than the capacity.
It is assumed that a required edge can be serviced from at least one of the depots in at least one direction.
The complexity of the initialization step changes from $\mc O(\lvert E_r\rvert)$ to $\mc O(\lvert V_d\rvert\lvert E_r\rvert)$.

\begin{procedure}[tbp]
	\small
	\LinesNumbered
	\SetKwInOut{Input}{Input}\SetKwInOut{Output}{Output}
	\SetKwRepeat{Do}{do}{while}%
	\SetKw{Continue}{continue}
	\SetKw{Break}{break}
	\SetKw{KwAnd}{and}
	\SetKw{KwOr}{or}
	\Input{$G=(V, E, E_r)$, depots $V_d$, costs, demands, capacity $Q$}
	\Output{Initialized coverage routes $\mc R$ with assigned depots}
	$\mc R\gets\emptyset$; $k \gets 1$\;
	\For(\tcp*[f]{$a_e$ and $\bar a_e$ are arcs for $e$}){$e \in E_r$}{
		$R_k\gets \emptyset$;\, $R_k.\opn{cost}\gets \infty$\;
		\For(\tcp*[f]{Iterate over depots}){$v \in V_d$}{
			$i\gets \opn{tail}(a_e)$;
			$j\gets \opn{head}(a_e)$\;
			$c \gets  \dcost{v, i} + \scost{a_e} + \dcost{j, v} + \lambda $\;
			$d \gets \ddemand{v, i} + \sdemand{a_e} + \ddemand{j, v}$\;
			\uIf{$d\leq Q$ \KwAnd $c < R_k.\opn{cost}$}{
				$R_k\gets a_e$;\, $R_k.v_0\gets v$\tcp*{Route with $v_0$}
			}
			$\bar c \gets \dcost{v, j} + \scost{\bar a_e} + \dcost{i, v} + \lambda $\;
			$\bar d \gets \ddemand{v, j} + \sdemand{\bar a_e} + \ddemand{i, v}$\;
			\uIf{$\bar d\leq Q$ \KwAnd $\bar c < R_k.\opn{cost}$}{
				$R_k\gets \bar a_e$;\, $R_k.v_0\gets v$\tcp*{Route with $v_0$}
			}
		}
		$\mc R.\operatorname{push}(R_k)$; $k\gets k+1$\;
	}
	\caption{MD-MEM::Initialize()}
	\label{alg:init_md}
\end{procedure}
\begin{procedure}[tbp]
	
		\small
		\LinesNumbered
		\SetKwInOut{Input}{Input}\SetKwInOut{Output}{Output}
		\SetKwRepeat{Do}{do}{while}%
		\SetKw{Continue}{continue}
		\SetKw{Break}{break}
		\SetKw{KwAnd}{and}
		\SetKw{KwOr}{or}
		\Input{Routes $\mc R_p$ and $\mc R_q$, $G=(V,E,E_r)$, depots $V_d$, capacity $Q$, costs, demands}
		\Output{Savings $s_{pq}$}
		$s_{pq}\gets -\infty$\;
		\ForEach{$v_0\in V_d$}{
			Compute saving $s_{v_0}$ for $R_p\mergeSym R_q$ and depot $v_0$\;
			\uIf{$s_{v_0} > s_{pq} $ \KwAnd $\operatorname{demand}(R_p\mergeSym R_q) \leq Q$} {
				$s_{pq}\gets s_{v_0}$\;
			}
		}
	\caption{MD-MEM::ComputeSavings()}
	\label{alg:savings_md}
\end{procedure}
\subsubsection{Integrated Multiple Depot Route Computation}
In the single depot version of the MEM algorithm, we considered eight permutations for merging two routes, four of which are shown in \fgref{fig:eightways}.
The cost saving for merging two routes $R_p$ and $R_q$ is given by $\opn{cost}(R_p) + \opn{cost}(R_q) - \opn{cost}(R_{pq})$, where $R_{pq}$ is the merged route.
In the multi-depot formulation, the cost and the demand of the merged route depend on the depot $v_0$---the merged route can be assigned to any of the depots, independent of the constituent routes.
However, the choice of the depot affects the savings in cost that can be achieved by merging two routes.
Thus, we iterate over all the depots for the multi-depot line coverage problem and check all eight merging permutations for each depot.
This gives us $8\lvert V_d\rvert$ computations, and the one with the maximum saving is selected provided it satisfies the capacity constraint.
The pseudocode for the computation of savings is given in Procedure~\ref{alg:savings_md}.

The most expensive component of the MEM algorithm is the embed step.
This step involves the computation of savings of a newly merged route with the others in the current set of routes~$\mc R$ (lines~\ref{alg:mem:embedstart}--\ref{alg:mem:embedend} in Algorithm~\ref{alg:mem}).
Thus, the overall complexity of the algorithm changes to $\mc O(\lvert V_d\rvert m^2 \log m)$.
As the computational cost depends on the number of depot locations, keeping their number as small as needed is advantageous.
In practice, making the number of depots equal to the number of robots available gives good results.

\subsection{Line Coverage with Turning Costs and Nonholonomic Constraints: MD-MEM-Turns}
\label{sec:turns}
Until now, we have considered the costs and the demands for traversing the edges to be arbitrary non-negative constants.
However, they do not model turning costs and nonholonomic constraints.
A smooth trajectory is desired in several robotics applications.
Even for differential drive robots, which can perform turns in place, a sharp turn is undesirable as the robot will need to slow down, take a turn, and then accelerate.
Smooth paths are often computed as a post-processing step after path planning~\cite{YangS10,RavankarRKHP18}.
However, such post-processing steps are performed individually for each route and do not account for overall cost minimization.
Furthermore, the resulting route can violate the constraints on the resource capacity of the robots.
Such path-smoothing techniques can be integrated with the MEM algorithm to minimize the total cost and ensure that the demand of a route is within the capacity.
This would require modification of the initialization procedure and the computation of savings, similar to the previous section for the multi-depot line coverage problem.
However, path smoothing can be an expensive process, and doing so $8\lvert V_d\rvert$ times for the embed step in each iteration may not be practical for applications that require rapid solutions.
This section instead illustrates constant-time procedures to generate routes with smooth turns and paths that respect nonholonomic constraints, and integrates them with the MEM algorithm.
Furthermore, the costs and the demands for taking turns are incorporated into the MEM algorithm.

\subsubsection{Smooth Turns and Turning Costs}
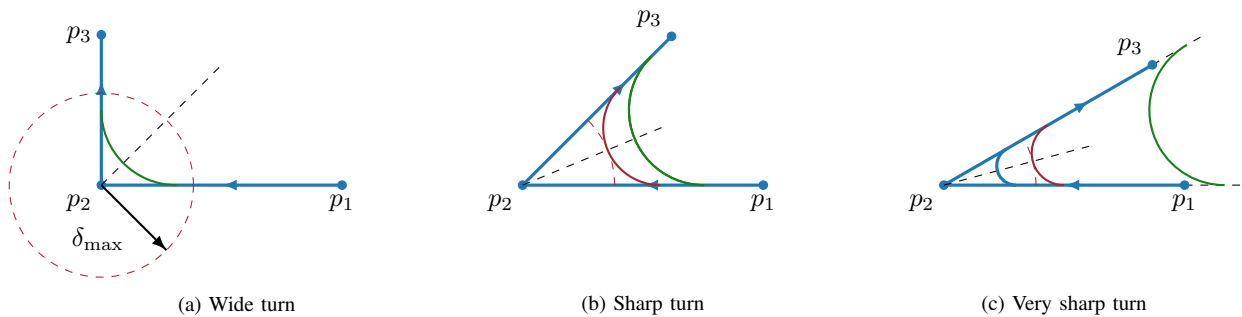
\begin{figure*}[htbp]
	\centering
	\begin{center}
	\begin{tikzpicture}[scale=0.4]
		\usetikzlibrary{calc}
		\tikzstyle{scEnd}=[near end, fill=none, mDarkRed]
		\tikzstyle{scSt}=[near start, fill=none, black]
		\tikzstyle{scMid}=[midway, fill=none, black]


		\coordinate (p1) at (8, 0);
		\coordinate (p2) at (0,0);
		\coordinate (p3) at (0, 5);

		\node[place]  at (p1)		{};
		\node[place]  at (p2)		{};
		\node[place]  at (p3)		{};
		\node[fill=none]  at (4.5, -4)		{\footnotesize (a) Wide turn};

		\node[fill=none, below]					at (p1)		{$p_1$};
		\node[fill=none, below left]		at (p2)		{$p_2$};
		\node[fill=none, left]					at (p3)		{$p_3$};

		\draw[mid2arc=0.5, mBlue, very thick] (p1) to (p2);
		\draw[mid2arc=0.7, mBlue, very thick] (p2) to (p3);

		\draw[thin, black, dashed] (p2) to (4, 4);
		\path[thin, black] (p2) edge [mid2arc=1.0] node[scMid,below left] {$\delta_{\max}$} (2.164, -2.164);

		\draw[thin, mDarkRed, dashed] (p2) circle [radius=3.06];
		\draw[thick, mGreen] (0,2.5)  arc[radius = 2.5, start angle= -180, end angle= -90];

		\coordinate (off) at (14,0);

		\coordinate (p1) at ($(off) + (p1)$);
		\coordinate (p2) at ($(off) + (p2)$);
		\coordinate (p3) at ($(off) + ({7*cos(45)},{7*sin(45)})$);

		\node[place]  at (p1)		{};
		\node[place]  at (p2)		{};
		\node[place]  at (p3)		{};

		\node[fill=none, below]					at (p1)		{$p_1$};
		\node[fill=none, below left]		at (p2)		{$p_2$};
		\node[fill=none, above left]					at (p3)		{$p_3$};

		\node[fill=none]  at (18, -4)		{\footnotesize (b) Sharp turn};
		\draw[mid2arc=0.5, mBlue, very thick] (p1) to (p2);
		\draw[mid2arc=0.7, mBlue, very thick] (p2) to (p3);

		\draw[thin, black, dashed] (p2) to ($(off) + ({5*cos(22.5)},{5*sin(22.5)})$);

		\centerarc[mDarkRed,thin,dashed](p2)(0:45:3.06);
		\centerarc[mGreen,thick]($(off) + (6.035,2.5)$)(-225:-90:2.5);
		\centerarc[mGreen,thick]($(off) + (6.035,2.5)$)(-225:-90:2.5);
		\centerarc[mDarkRed,thick]($(off) + (4.58,1.90)$)(-225:-90:1.90);

		\coordinate (p1) at ($(off) + (p1)$);
		\coordinate (p2) at ($(off) + (p2)$);
		\coordinate (p3) at ($(p2) + ({8*cos(30)},{8*sin(30)})$);
		\draw[thin, black, dashed] (p3) to ($(p2) + ({10*cos(30)},{10*sin(30)})$);
		\draw[thin, black, dashed] (p1) to ($(p2) + (10,0)$);

		\node[place]  at (p1)		{};
		\node[place]  at (p2)		{};
		\node[place]  at (p3)		{};

		\node[fill=none, below]					at (p1)		{$p_1$};
		\node[fill=none, below left]		at (p2)		{$p_2$};
		\node[fill=none, above left]					at (p3)		{$p_3$};

		\node[fill=none]  at (32, -4)		{\footnotesize (c) Very sharp turn};
		\draw[mid2arc=0.5, mBlue, very thick] (p1) to (p2);
		\draw[mid2arc=0.7, mBlue, very thick] (p2) to (p3);

		\draw[thin, black, dashed] (p2) to ($(p2) + ({5*cos(15)},{5*sin(15)})$);

		\centerarc[mDarkRed,thin,dashed](p2)(0:30:3.06);
		\centerarc[mDarkRed,thick]($(p2) + (3.987,1.06855)$)(-240:-90:1.06855);
		\centerarc[mGreen,thick]($(p2) + (9.33,2.5)$)(-240:-90:2.5);
		\centerarc[mBlue,very thick]($(p2) + (2.401,0.643)$)(-240:-90:0.643);
	\end{tikzpicture}
\end{center}
	\caption[Visualization of smooth turns for adjacent required edges]{Smooth turns for a robot traversing two adjacent edges from $p_1$ to $p_3$ through $p_2$.
		The minimum turning radius is given by the ratio of its linear and angular velocities, and the corresponding arc is shown in green.
		(a)~When the corner is wide, the robot can turn smoothly without changing its velocity and deviating too much from the original path.
		(b)~However, if the turn is sharp, the robot needs to slow down so that the deviation is within the permitted limit $\delta_{\max}$.
		The optimal turning arc is shown in red.
		(c)~If the turn is very sharp, the robot may not have enough distance to decelerate to the required velocity for the red arc.
		In the worst case of a 180-degree turn, it may have to come to a complete stop.
		The innermost blue arc shows the optimal turning arc based on the maximum deceleration and the turning angle.
	\label{fig:smooth_turns}}
\end{figure*}

We first consider the generation of routes with smooth turns for holonomic robots.
We will later see that such smooth turns are useful for nonholonomic robots as well.
For a specified linear velocity when a robot is in motion, the minimum turning radius for a robot is given by the ratio of its linear and angular velocities; if the turning radius is small, the robot can take sharper turns.
When two adjacent edges do not have a sharp corner, the robot can take a smooth turn without deviating too much from the edges, as shown by the green arc in \fgref{fig:smooth_turns}(a).
The arc is tangential to the two edges.
A user-defined parameter $\delta_{\max}$ determines the maximum allowable deviation from the edges, as shown by the red dashed circle.
The parameter can be set based on the sensor field-of-view to ensure coverage of every point on the corresponding linear features.
However, if the turn is sharp, the robot needs to slow down so that the maximum deviation is within $\delta_{\max}$, as shown by the red arc in \fgref{fig:smooth_turns}(b).
This requires the robot to decelerate to achieve the required turning radius by decreasing the linear velocity.
When a turn is very sharp, as shown in \fgref{fig:smooth_turns}(c), the robot may not have enough length available to decelerate to the required velocity for the red arc corresponding to $\delta_{\max}$.
This requires solving a quadratic equation to determine a time-optimal turning arc, which corresponds to the innermost blue arc in the figure.
In the worst case of a 180-degree turn, the robot may need to come to a complete stop.

The primary computation step in the MEM algorithm is the computation of savings.
	Consider two routes $R_p$ and $R_q$ with end edges $p$ and $q$, respectively, for a particular merge permutation.
	The algorithm computes the turning cost to go from edge $p$ to edge $q$.
	However, such a turn may be sharp and require the robot to decelerate.
	This, in turn, can affect the velocities on edges $p-1$ and $q-1$ in their respective routes, potentially resulting in a cascading effect that may require computations proportional to the number of edges.
	To avoid this, we enforce that the robot can decelerate only after it has reached the middle of the first edge~$p$ and can accelerate until the middle of the next edge~$q$.
	By avoiding a cascading effect, this modularizes the computation of the turning costs and ensures constant time computation of savings.
	As the accelerations of the robots are usually high, and the lengths of the edges are comparatively large in practical applications, we can assume that the robot can come to a complete stop from its full speed within half the length of the edge; otherwise, the robot may not be able to take 180-degree turns.


\begin{figure*}[ht]
	\centering
	\subfloat[No adjacent edges, Dubins curves]{%
	\includegraphics[height=0.16\textheight,trim={1.2cm 0cm 0cm 0cm},clip]{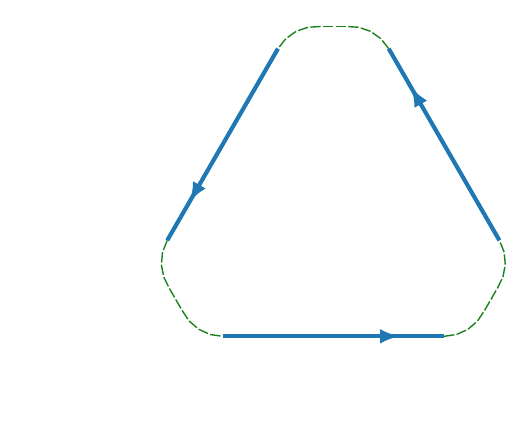}}
	\hfill
	\subfloat[Adjacent edges, Dubins curves]{%
	\includegraphics[height=0.16\textheight]{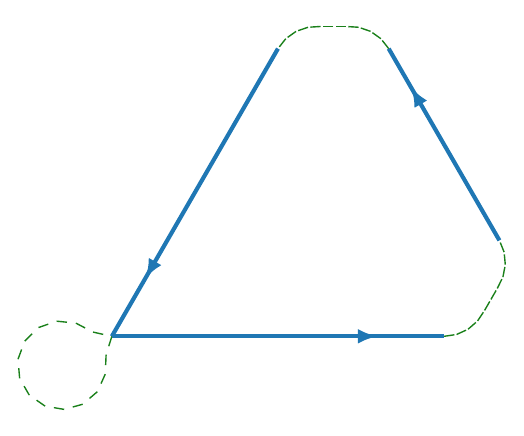}}
	\hfill
	\subfloat[Adjacent edges, Dubins curves and smooth turns]{%
	\includegraphics[height=0.16\textheight,trim={1.0cm 0cm 0cm 0cm},clip]{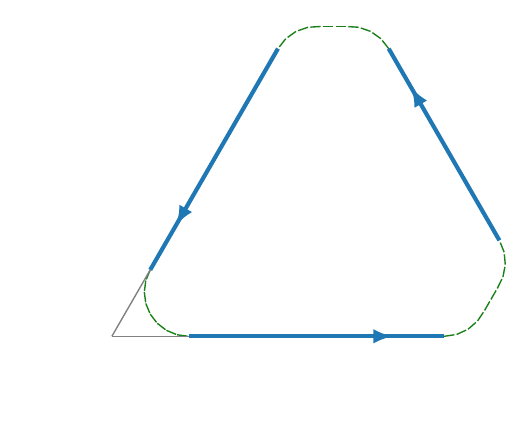}}
	\caption[Modeling of deadheading paths for nonholonomic robots]{%
		Modeling of deadheading paths for nonholonomic robots.
		(a)~The graph does not contain sharp turns, and Dubins curves give optimal paths for deadheading.
		(b)~When we have adjacent edges, common in road networks, Dubins curves can create several circular arcs to orient the robot along the edges.
		(c)~We introduce smooth turns with Dubins curves to generate efficient deadheading paths between required edges.
	\label{fig:dubins_triangle}}
\end{figure*}


\subsubsection{Nonholonomic Robots}
Several commonly used robots, such as car-like robots, fixed-wing UAVs, and underwater robots, have nonholonomic constraints and cannot take turns in place.
We use a unicycle model of the robots~\cite{LynchParkbook} and incorporate Dubins curves for the motion of the robots.
Dubins curves are often used to determine optimal paths from one pose of the robot to the other~\cite{Lavalle06}, where the pose comprises the position and the orientation of the robot.
An example of a coverage route using Dubins curves to determine optimal deadheadings is given in \fgref{fig:dubins_triangle}(a).
A Reeds-Shepp model can also be used to determine such turns.

The Dubins curves can be inefficient when we have adjacent required edges as we may have to take extra turns to align the heading of the robot with the subsequent edge, as shown in \fgref{fig:dubins_triangle}(b).
Instead, we leverage the smooth turn model, described in the previous subsection, to allow the robots to deviate from their path within a given limit $\delta_{\max}$.
This eliminates most of the extra turns generated by just using Dubins curves.
An example is shown in \fgref{fig:dubins_triangle}(c).
In the case of fixed-wing UAVs that have a lower bound on the minimum speed, the algorithm selects Dubins curves for very sharp turns.

\subsubsection{The MD-MEM-Turns Algorithm}
Given an initial and a final heading angle at the depots, we require three kinds of cost functions for deadheading:
(1)~from a depot to any required arc $c(v_0, a)$,
(2)~from one required arc to another $c(a_1, a_2)$, and
(3)~from a required arc to a depot $c(a, v_0)$.
Here, $v_0$ is a depot, and $a, a_1, a_2 \in \mc A_r$ are required arcs.
For environments with cost functions that satisfy the triangle inequality, these cost functions can be computed in constant time using the above procedures for smooth turns and Dubins curves.
When the triangle inequality is not satisfied, we use the technique of line graphs\footnote{\cite{Winter02,GeisbergerV11} refer to line graphs as dual graphs, which should not be confused with dual graphs in graph theory.}~\cite{Winter02,GeisbergerV11} to compute the shortest deadheading paths and cost functions, which can be computed in $\mc O(\lvert E\rvert^3)$ time, where $E$ is the number of edges in the graph.
The procedure~\ref{alg:init_md_turns} shows the modified initialization using multiple depots and turning costs.
Similarly, the cost savings for merging can be modified to incorporate turning costs.
The running time of the MEM algorithm does not change and is given by $\mathcal O(\lvert V_d\rvert m^2 \log m)$.
\begin{procedure}[tbp]
	\small
	\LinesNumbered
	\SetKwInOut{Input}{Input}\SetKwInOut{Output}{Output}
	\SetKwRepeat{Do}{do}{while}%
	\SetKw{Continue}{continue}
	\SetKw{Break}{break}
	\SetKw{KwAnd}{and}
	\SetKw{KwOr}{or}
	\Input{$G=(V, E, E_r)$, depots $V_d$, costs, demands, capacity $Q$}
	\Output{Initialized coverage routes $\mc R$ with assigned depots}
	$\mc R\gets\emptyset$; $k \gets 1$\;
	\For(\tcp*[f]{$a_e$ and $\bar a_e$ are arcs for $e$}){$e \in E_r$}{
		$R_k\gets \emptyset$;\, $R_k.\opn{cost}\gets \infty$\;
		\For(\tcp*[f]{Iterate over depots}){$v \in V_d$}{
			$c \gets  \dcost{v, a_e} + \scost{a_e} + \dcost{a_e, v} + \lambda $\;
			$d \gets \ddemand{v, a_e} + \sdemand{a_e} + \ddemand{a_e, v}$\;
			\uIf{$d\leq Q$ \KwAnd $c < R_k.\opn{cost}$}{
				$R_k\gets a_e$;\, $R_k.v_0\gets v$\tcp*{Route with $v_0$}
			}
			$\bar c \gets \dcost{v, \bar a_e} + \scost{\bar a_e} + \dcost{\bar a_e, v} + \lambda $\;
			$\bar d \gets \ddemand{v, \bar a_e} + \sdemand{\bar a_e} + \ddemand{\bar a_e, v}$\;
			\uIf{$\bar d\leq Q$ \KwAnd $\bar c < R_k.\opn{cost}$}{
				$R_k\gets \bar a_e$;\, $R_k.v_0\gets v$\tcp*{Route with $v_0$}
			}
		}
		$\mc R.\operatorname{push}(R_k)$; $k\gets k+1$\;
	}
	\caption{MD-MEM-Turns::Initialize()}
	\label{alg:init_md_turns}
\end{procedure}

\section{Simulations and Experiments}
\label{sc:sim}
This section empirically demonstrates the high-quality solutions provided by the Merge-Embed-Merge (MEM) algorithm for the line coverage problem with multiple resource-constrained robots.
We analyze the performance of the MEM algorithm in terms of computation time and the solution quality on a dataset of 50 road networks.
We then demonstrate extensions of the MEM algorithm for large-scale graphs using a multi-depot formulation and for nonholonomic robots.
We perform experiments with aerial robots on the UNC Charlotte campus road network.
Finally, we demonstrate how the line coverage problem can be applied to the area coverage problem. 


We use the DJI Phantom~4 quadrotor in our experiments, with the following cost model for traversing the edges.
Denote the speed of the UAV by~$v$ and the wind speed by~$w$.
Let the travel vector~$\mv t$ denote the traversal of an edge from its tail vertex $v_t$ to the head vertex $v_h$.
Let $\phi$ be the angle between the wind vector and the travel vector~$\mv t$ for an edge.
Then the effective speed of the UAV is given by:
\begin{align*}
	v_{\text{eff}} = w \cos{\phi} + \sqrt{v^2 - w^2 \sin^2{\phi}}
\end{align*}
The cost function is defined as the time taken for the UAV to traverse an edge:
\begin{align*}
	c(v_t, v_h) = \frac{\norm{\mv t}_2}{v_\text{eff}}
\end{align*}
Here, $\norm{\mv t}_2$ is the Euclidean distance from $v_t$ to $v_h$.
We use different speeds for servicing and deadheading, and the velocity~$v$ is set accordingly based on the travel mode.
Note that the cost function is direction-dependent due to wind, and hence, the graph is asymmetric.

For convenience, we use travel time as the cost and demand functions and specify the capacity in terms of the allowable flight time for the UAVs.
These functions need not be the same in practice; instead, functions that model battery consumption can be used instead.
Our open-source implementation allows for any non-negative edge cost and demand functions.

\begin{figure*}[htbp]
	\centering
	\includegraphics[width=0.95\textwidth]{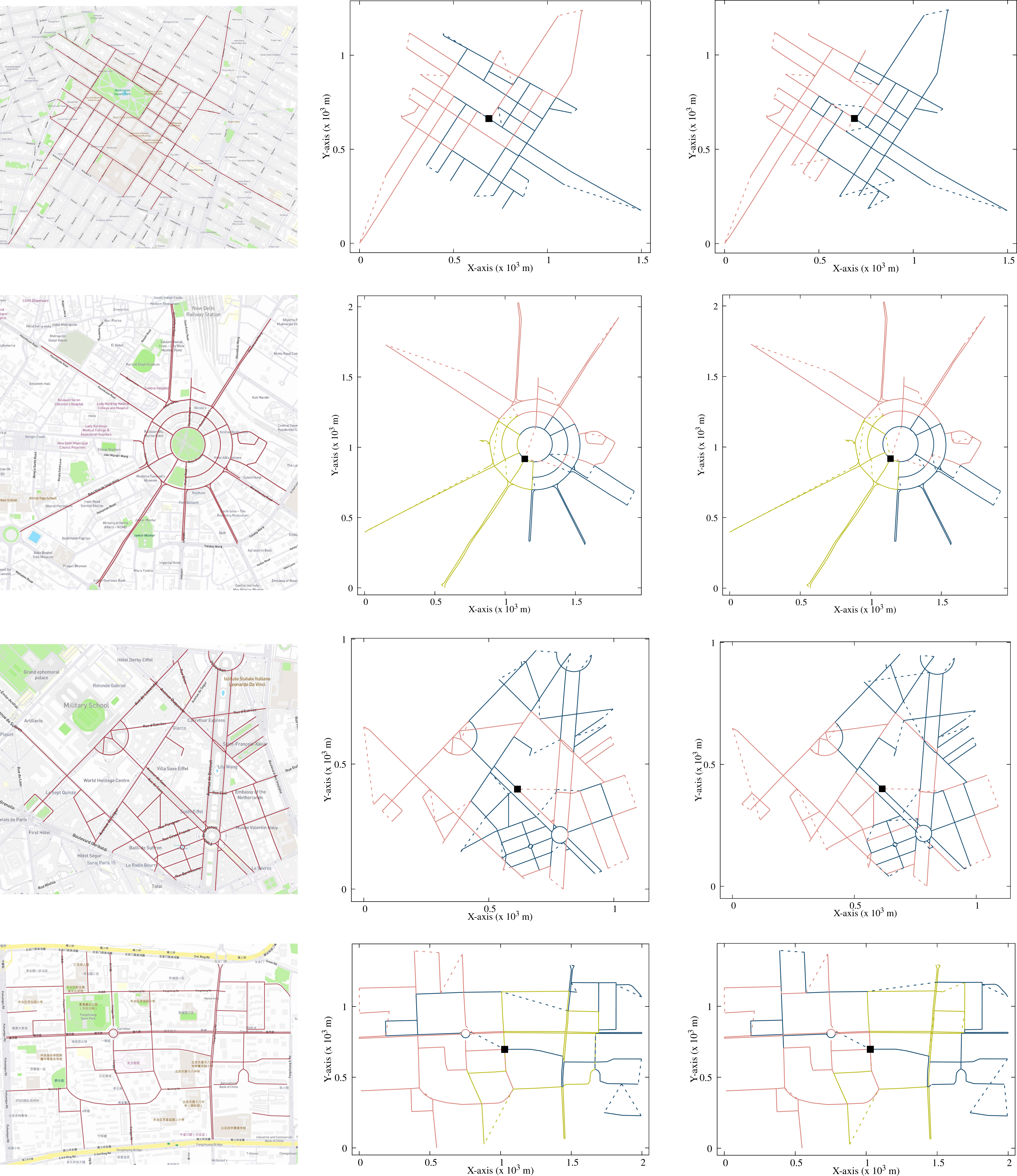}
	\caption[Four sample road networks with ILP and MEM routes]{Four of the 50 sample road networks obtained for the most populous cities:
		The first column is the map with the input graph, the second column is the solution obtained using the ILP formulation, and the third column is the solution obtained using the MEM algorithm.
		The road networks, from top to bottom, are from (a)~New York, (b)~Delhi, (c)~Paris, and (d)~Beijing.
		Only the required edges are shown in the input graph, and there is a required edge for each pair of vertices in the graph.
		For example, the New York graph has 379 vertices, 402 required edges, and 71,361 non-required edges.
		The solid lines represent servicing travel mode, and the dashed lines represent deadheading travel mode.
	\label{fig:sample}}
\end{figure*}

\subsection{Analysis on Road Networks with Single Depot}
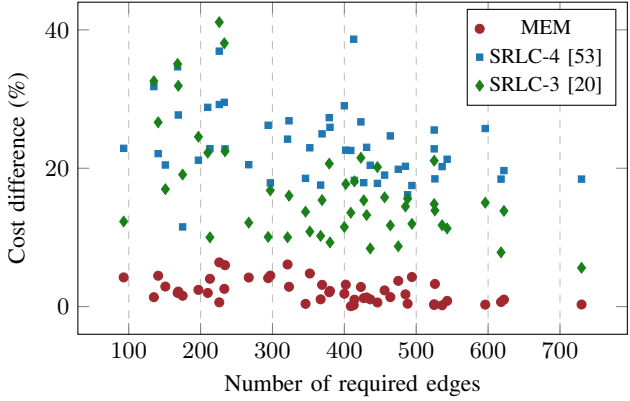
\begin{figure}[t]
	\centering
	\begin{center}
\begin{tikzpicture}
	\small
	\begin{axis}[
		enlargelimits=true,
		xlabel = {Number of required edges},
		ylabel = {Cost difference (\%)},
		xmajorgrids=true,
		grid style=dashed,
		legend style={font=\footnotesize},
		xtick={0,100,200,300,400,500,600,700,800},
		height=6cm,
		ymin=0,
		ymax=40,
		ylabel near ticks
		]
		\addplot[
		only marks,
		color=mDarkRed,
		mark=*,
		mark size=1.7pt]
		table[x=m,y=mem, col sep=comma]
		{./graphics/data/mem_ilp_b3a_1200.csv};
		\addplot+[
		only marks,
		color=mBlue,
		mark options={fill=mBlue},
		mark=square*,
		mark size=1.2pt]
		table[x=m,y=b3a, col sep=comma]
		{./graphics/data/mem_ilp_b3a_1200.csv};
		\addplot+[
		only marks,
		color=mGreen,
		mark options={fill=mGreen},
		mark=diamond*,
		mark size=2.0pt]
		table[x=m,y=b2a, col sep=comma]
		{./graphics/data/mem_ilp_b3a_1200.csv};
	\legend{MEM, SRLC-4~\cite{vanBevernKS17}, SRLC-3~\cite{AgarwalA20WAFR}}
	\end{axis}
\end{tikzpicture}
\end{center}
	\caption[Comparison of solutions generated using MEM and ILP]{Comparison of line coverage algorithms, computed as a cost difference percentage of the ILP solution.
		The MEM solutions, shown as red circles, are within 7\% of the ILP solutions, and are significantly better than the 4-approx.~\cite{vanBevernKS17} and the 3-approx.~\cite{AgarwalA20WAFR} algorithms for single robot line coverage without capacity constraints.
The number of required edges corresponds to the number of line segments in the road networks.\label{fig:cost1200}}
\end{figure}

We first perform a detailed simulation analysis for the single-depot line coverage problem to establish the efficacy and efficiency of the MEM algorithm.
The analysis is performed on a dataset of 50 road networks, spanning about 1\SIkmsqr{} area, from the most populous cities around the world, which we created using OpenStreetMap~\cite{OpenStreetMap}.
The road networks are representative of environments with linear features, and they provide widely varying graph structures for a thorough evaluation of the algorithm.
These networks are represented by line segments, which form the set of required edges.
The endpoints and the intersections of the road segments form the vertex set in the graph.
For UAVs that fly at high altitudes, we add a non-required edge between each pair of vertices if a required edge does not exist between them.

The MEM algorithm is implemented in C++ and executed on a standard laptop with an Intel Core i7-1195G7 processor on a single core.
We compare the quality of the solutions computed using the MEM algorithm with the solutions from solving the ILP formulation.
The ILP formulation is solved using Gurobi~\cite{Gurobi}, and the C++ API was used to interface with the solver.
The ILP formulation, when executed on the laptop for an hour, is unable to compute a feasible solution for 41 of the 50 road networks.
Thus, it is not very practical as the environment conditions can change, and efficient solutions may be needed quickly in response to emergency situations.
Nevertheless, the ILP formulation provides an excellent benchmark to evaluate the quality of the heuristic solutions.
We use the solutions obtained using the MEM algorithm to provide an initial solution to the solver, which helps in upper-bounding the branch-and-bound algorithm used by ILP solvers.
Thus, the ILP solutions are always at least as good as the initial solutions.
To obtain high-quality solutions, the ILP formulation is executed on a cluster node with an Intel Xeon Gold 6248R processor using 16 cores in parallel for each road network, and the execution time is limited to 24~hours.

We set the servicing and the deadheading speeds to 7\SIvel{} and 10\SIvel{}, respectively.
A wind of 2\SIvel{} is simulated from the southwest direction, i.e., $\pi/4$ radians from the horizontal axis.
For the first set of experiments, we set the capacity of the robot to 20 minutes (1,200\SIs{}).
\fgref{fig:sample} shows four of the fifty road networks, along with the routes obtained using the ILP formulation and the MEM algorithm.
We consider a single depot location for these simulations, shown by a black square.
The depot is set to be the vertex closest to the mean of all the vertices in the graph.
The road networks provide a rich set of graph structures, as illustrated by the four sample graphs.
The MEM solutions look similar to that of the ILP formulation, and the amount of deadheading is minimal.

\fgref{fig:cost1200} presents the performance of the MEM algorithm, shown as red circles, in terms of cost difference percentage, computed as $100\frac{(c-c^*)}{c^*}$, where $c$ and $c^*$ are the costs of solutions computed using the algorithm and the ILP formulation, respectively.
The solutions are compared with ones computed using the 4-approximation algorithm~\cite{vanBevernKS17} and the 3-approximation algorithm~\cite{AgarwalA20WAFR} for the single robot line coverage.
Even though the algorithms for the single robot case have theoretical bounds on the cost, they do not perform as well as the MEM algorithm for the road network dataset.
Note that the single robot algorithms are solving a much simpler problem as they do not consider resource demands, capacity constraints, and depot locations.
The difference in cost of the MEM solutions does not deviate significantly as the number of required edges increases.
The performance of the ILP decreases rapidly for large graphs as the number of variables increases, and the ILP is not able to converge to an optimal solution within the computation time of 24~hours.
Furthermore, graphs with a large network length require more routes, increasing the number of variables in the ILP formulation.

\begin{figure}[th!]
	\centering
\begin{center}
	\begin{tikzpicture}
		\small
		\pgfplotsset{
			my boxplot style/.style={
				boxplot,
				draw=black,
				solid,
				fill=white,
				mark=*,
				every mark/.append style={
					fill=white,
					draw=black,
				},
			},
		}
		\makeatletter
		\pgfplotsset{
			boxplot/draw/average/.code={%
				\color{black}          
				\draw[/pgfplots/boxplot/every average/.try]
				\pgfextra
				\pgftransformshift{%
					\pgfplotsboxplotpointabbox
					{\pgfplotsboxplotvalue{average}}
					{0.5}%
				}%
				\pgfuseplotmark{\tikz@plot@mark}%
				\endpgfextra
				;
			},
		}
		\makeatother
		\begin{axis}[
			enlargelimits=true,
			boxplot,
			boxplot/draw direction=y,
			xtick={1,2,3,4,5,6,7},
			xticklabels={100,200,300,400,500,600,700},
			boxplot/average={auto},
			height=6cm,
			ylabel = {Computation time (s)},
			xlabel = {Number of required edges},
			ylabel near ticks,
			boxplot={
				draw position={1 + floor(\plotnumofactualtype/2) + floor(\plotnumofactualtype/12) - floor(\plotnumofactualtype/13)},
				box extend=0.5
			},
			]
			\addplot[my boxplot style,draw=mDarkRed]table[y=t, col sep=comma]{./graphics/data/timings_100.csv};
			\addplot+[my boxplot style,draw=mBlue]table[y=m, col sep=comma]{./graphics/data/timings_100.csv};
			\addplot+[my boxplot style,draw=mDarkRed]table[y=t, col sep=comma]{./graphics/data/timings_200.csv};
			\addplot+[my boxplot style,draw=mBlue]table[y=m, col sep=comma]{./graphics/data/timings_200.csv};
			\addplot+[my boxplot style,draw=mDarkRed]table[y=t, col sep=comma]{./graphics/data/timings_300.csv};
			\addplot+[my boxplot style,draw=mBlue]table[y=m, col sep=comma]{./graphics/data/timings_300.csv};
			\addplot+[my boxplot style,draw=mDarkRed]table[y=t, col sep=comma]{./graphics/data/timings_400.csv};
			\addplot+[my boxplot style,draw=mBlue]table[y=m, col sep=comma]{./graphics/data/timings_400.csv};
			\addplot+[my boxplot style,draw=mDarkRed]table[y=t, col sep=comma]{./graphics/data/timings_500.csv};
			\addplot+[my boxplot style,draw=mBlue]table[y=m, col sep=comma]{./graphics/data/timings_500.csv};
			\addplot+[my boxplot style,draw=mDarkRed]table[y=t, col sep=comma]{./graphics/data/timings_600.csv};
			\addplot+[my boxplot style,draw=mBlue]table[y=m, col sep=comma]{./graphics/data/timings_600.csv};
			\addplot+[my boxplot style,draw=mDarkRed]table[y=t, col sep=comma]{./graphics/data/timings_700.csv};
			\addplot+[my boxplot style,draw=mBlue]table[y=m, col sep=comma]{./graphics/data/timings_700.csv};
		\end{axis}
	\end{tikzpicture}
\end{center}
	\caption[Computation time for the MEM algorithm]{Computation time for the MEM algorithm on the road network dataset.
		Each road network is placed in bins of size 100 based on the number of required edges.
		The MEM algorithm is executed 100 times for each road network.
		The results are shown as a boxplot: the red boxes represent the total computation time, while the blue boxes represent the time taken by the MEM algorithm.
		Diamond markers show the average, and the circles are outliers.%
	\label{fig:timings}}
\end{figure}
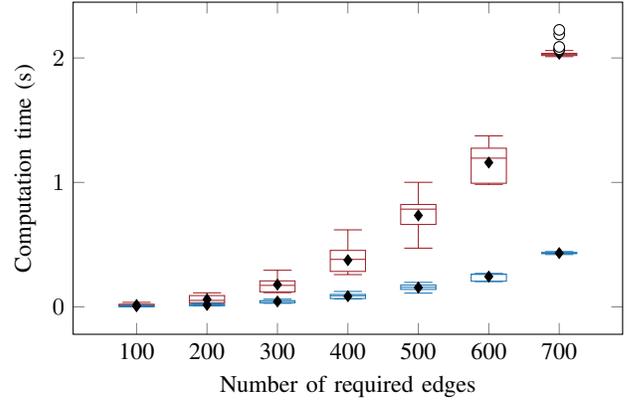

\fgref{fig:timings} shows the computation time required to obtain solutions for the road network dataset.
Most of the computation is spent processing the network and creating the graph.
The largest road network with 730 required edges is solved in about 2\SIs{}, of which the MEM algorithm takes less than 0.5\SIs{}.
For graphs with less than 200 required edges, the solutions are generated within 0.1\SIs{}.
The results indicate that the algorithm is very fast for robotics applications.
It can also be used in real-time, where the graph structure and the remaining battery life are updated as robots traverse the environment.

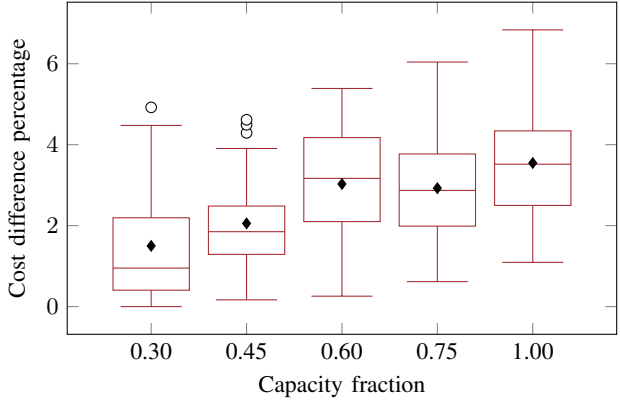
\begin{figure}[t!]
	\centering
	\usepgfplotslibrary{statistics}
\begin{center}
	\begin{tikzpicture}
		\small
		\pgfplotsset{
			my boxplot style/.style={
				boxplot,
				draw=black,
				solid,
				fill=white,
				mark=*,
				every mark/.append style={
					fill=white,
					draw=black,
				},
			},
		}
		\makeatletter
		\pgfplotsset{
			boxplot/draw/average/.code={%
				\color{black}          
				\draw[/pgfplots/boxplot/every average/.try]
					\pgfextra
					\pgftransformshift{%
						\pgfplotsboxplotpointabbox
						{\pgfplotsboxplotvalue{average}}
						{0.5}%
					}%
					\pgfuseplotmark{\tikz@plot@mark}%
					\endpgfextra
					;
			},
		}
		\makeatother
		\centering
		\pgfplotstableread[col sep=comma]{./graphics/cost_diff.csv}{\costdata};
	\begin{axis}[
		boxplot,
		boxplot/draw direction=y,
		xtick={1,2,3,4,5},
		ytick={0,2,4,6,8},
		xticklabels={0.30,0.45,0.60,0.75,1.00},
		enlarge y limits,
		height=6cm,
		legend style={font=\footnotesize},
		boxplot/average={auto},
		ylabel = {Cost difference percentage},
		xlabel = {Capacity fraction},
		]
		\addplot+[my boxplot style,draw=mDarkRed]table[y=Cost_diff30]{\costdata};
		\addplot+[my boxplot style,draw=mDarkRed]table[y=Cost_diff45]{\costdata};
		\addplot+[my boxplot style,draw=mDarkRed]table[y=Cost_diff60]{\costdata};
		\addplot+[my boxplot style,draw=mDarkRed]table[y=Cost_diff75]{\costdata};
		\addplot+[my boxplot style,draw=mDarkRed]table[y=Cost_diffInf]{\costdata};
	\end{axis}
\end{tikzpicture}
\end{center}
	\caption[Variation of cost with capacities]{%
		The cost difference between the solutions generated using the MEM algorithm and the ILP solution for different capacities.
		For each instance, the capacity is set as a fraction of the route cost for a single robot with infinite capacity.
		 In the boxplots, circles show the outliers, and the diamond markers show the average.
		The average difference in the cost between the solutions generated by the MEM algorithm and the ILP formulation is 2.61\%.%
	\label{fig:boxplot}}
\end{figure}

Next, we analyze the performance of the MEM algorithm with different robot capacities.
For each road network, the capacity is set as a fraction of the minimum cost required to cover the entire network using a single robot.
\fgref{fig:boxplot} shows the cost difference percentage between the solutions obtained using the ILP formulation and the MEM algorithm with varying capacity fractions.
Observe that the cost difference percentage decreases as the capacity decreases, i.e., the performance of the MEM algorithm with respect to the ILP formulation improves.
This happens because the number of routes~$K$ required to cover the entire road network increases as the capacity decreases, which, in turn, increases the number of variables in the ILP formulation.
As a result, the ILP formulation is not able to compute near-optimal solutions within the computation time limit of 24~hours.
In contrast, the performance of the MEM algorithm with respect to the ILP formulation, as given by the cost difference percentage, remains consistent for different capacities.
Furthermore, the running time of the MEM algorithm is not significantly affected as it requires fewer merges as the capacity decreases.

The simulations on the road network dataset show that the MEM algorithm generates solutions with costs comparable to that of the bounded-computation-time ILP formulation, i.e., within~7\%, and with a maximum computation time of around~2\SIs{}.
These results show that the algorithm is suitable for deploying robots, in particular aerial robots, for coverage of linear infrastructure.

\subsection{Analysis on 50 Large Road Networks with Multiple Depots}
We present simulation analysis for 50 large road networks, each spanning a 3\SIkm{}$\,\times\,$3\SIkm{} area.
The number of required edges ranges from 481 to 4831, and the network length from 22\SIkm{} to 108\SIkm{}.
The robot and environment parameters are the same as in the previous section.
Since the road networks are large, it is not possible to service the entire network from a single depot.
Thus, we use the multi-depot approach to generate line coverage routes, as discussed in \scref{sec:mdmem}.
Depot locations are generally selected based on the ease of field operation of robots.
In our simulations, the depot locations are randomly selected from the vertices of the required edges of the road network.
The number of depots is set based on the total average demand as follows:
\begin{equation}
	\label{eq:depots}
	\text{number of depots} = 1+\left\lceil\sum_{e\in E_r}\frac{\sdemand{a_e} + \sdemand{\bar a_e}}{2 Q}\right\rceil.
\end{equation}

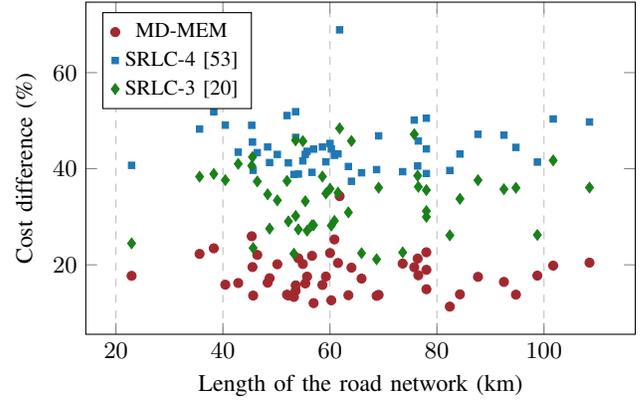
\begin{figure}[t!]
	\centering
	\begin{center}
\begin{tikzpicture}
	\small
	\begin{axis}[
		enlargelimits=true,
		xlabel = {Length of the road network (km)},
		ylabel = {Cost difference (\%)},
		xmajorgrids=true,
		grid style=dashed,
		legend style={font=\footnotesize},
		legend pos=north west,
		height=6cm,
		ylabel near ticks
		]
		\addplot[
		only marks,
		color=mDarkRed,
		mark=*,
		mark size=1.7pt]
		table[x expr=\thisrow{length_m}/1000,y=mem_p, col sep=comma]
		{./graphics/data/costs_large.csv};
		\addplot+[
		only marks,
		color=mBlue,
		mark options={fill=mBlue},
		mark=square*,
		mark size=1.2pt]
		table[x expr=\thisrow{length_m}/1000,y=b3_p, col sep=comma]
		{./graphics/data/costs_large.csv};
		\addplot+[
		only marks,
		color=mGreen,
		mark options={fill=mGreen},
		mark=diamond*,
		mark size=2.0pt]
		table[x expr=\thisrow{length_m}/1000,y=b2_p, col sep=comma]
		{./graphics/data/costs_large.csv};
	\legend{MD-MEM, SRLC-4~\cite{vanBevernKS17}, SRLC-3~\cite{AgarwalA20WAFR}}
	\end{axis}
\end{tikzpicture}
\end{center}
	\caption{
		Performance analysis of the MD-MEM algorithm for large road networks spanning 3\SIkm{}$\,\times\,$3\SIkm{} area with multiple depots.
	The MD-MEM solutions, shown as red circles, are consistently better than the 4-approx.~\cite{vanBevernKS17} and the 3-approx.~\cite{AgarwalA20WAFR} algorithms for single robot line coverage without capacity constraints.}
	\label{fig:costs_large}
\end{figure}

\begin{figure*}[htbp]
	\centering
	\subfloat[London]{%
		\hspace{0.5cm}
	\includegraphics[height=0.22\textheight,trim={0cm 0cm 0cm 0cm},clip]{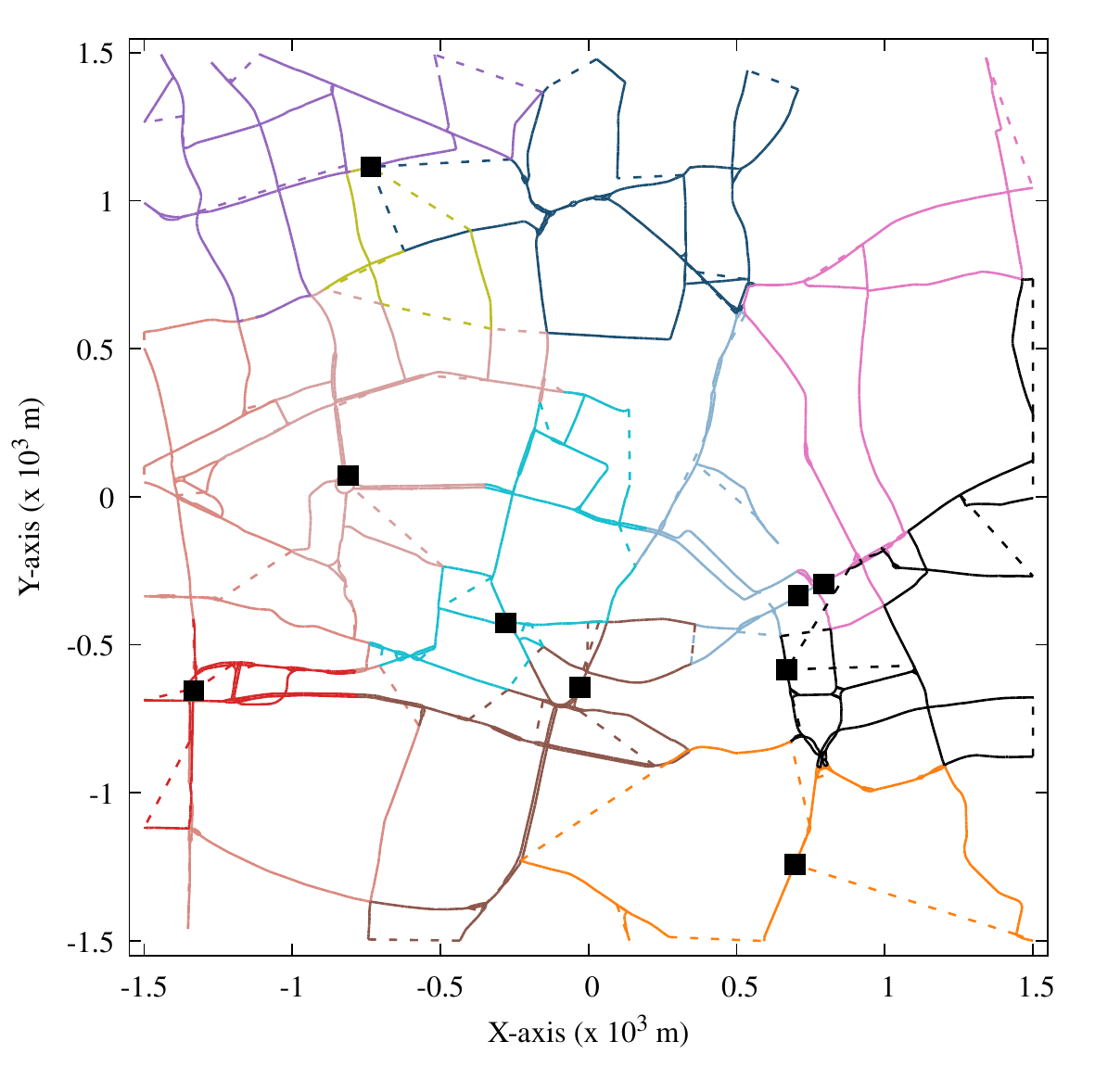}}
	\hfill
	\subfloat[Kolkata]{%
	\includegraphics[height=0.22\textheight]{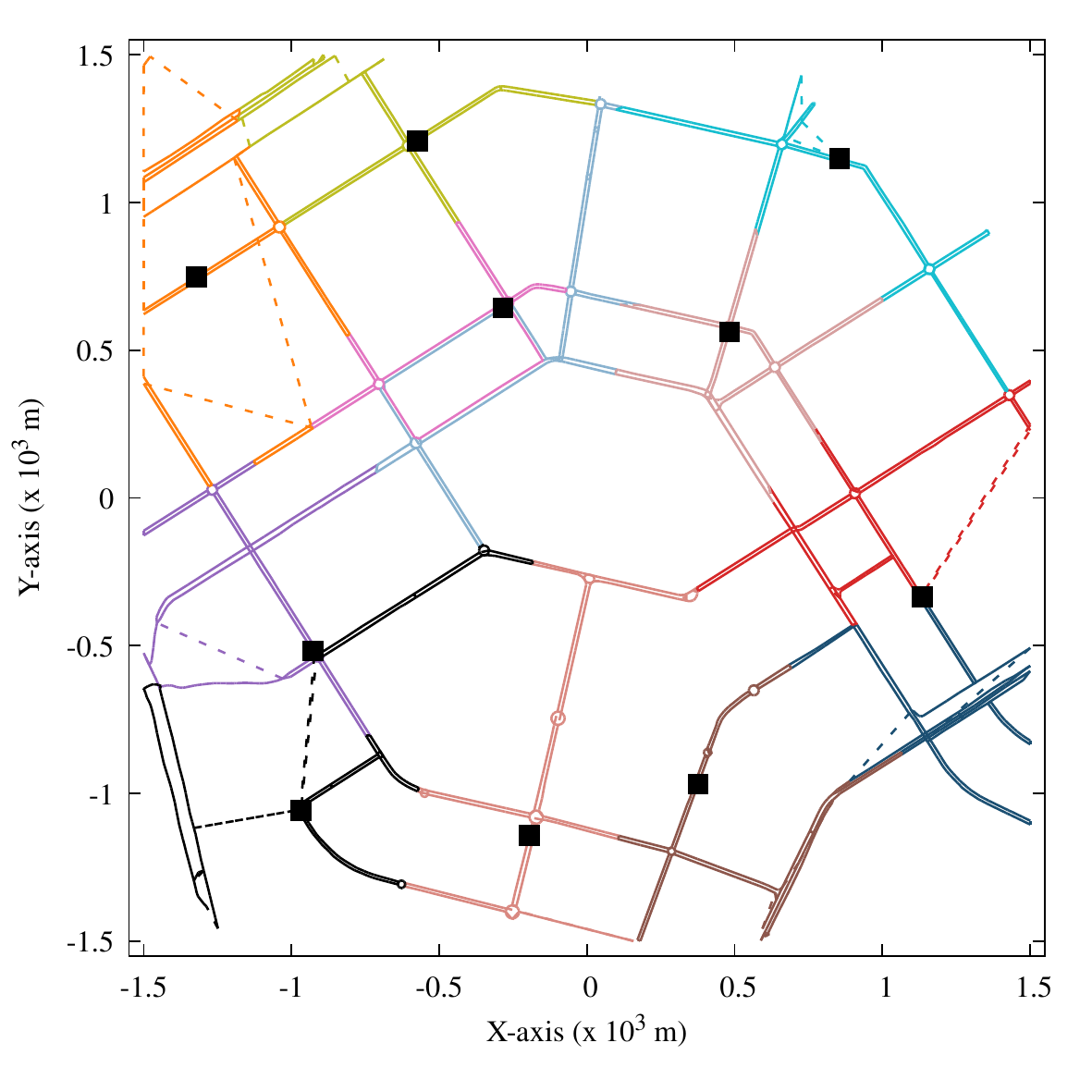}}
	\hfill
	\subfloat[Lima]{%
	\includegraphics[height=0.22\textheight]{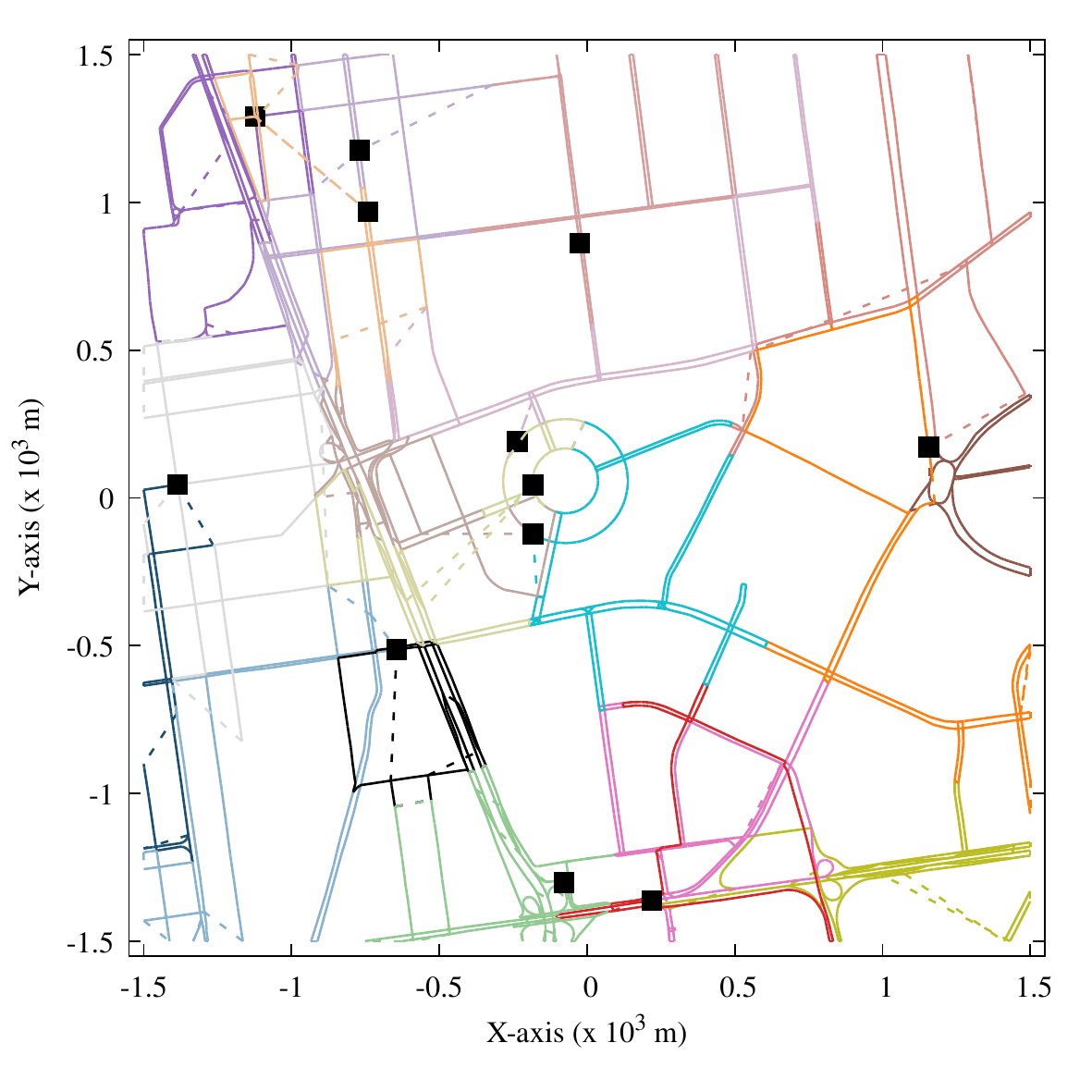}}
	\hspace{0.5cm}
	\caption[MD-MEM solutions for large dataset]{%
		Line coverage solutions for large road networks, each spanning a 3\SIkm{}$\,\times\,$3\SIkm{} area.
		The solutions are generated using the multi-depot merge-embed-merge (MD-MEM) algorithm.
		(a)~The London road network has the largest number of required edges in the dataset, with 4831 required edges, over 10 million non-required edges, and about 60~\SIkm{} total network length.
		(b)~The Kolkata road network has 2160 required edges and has a total network length of 66~\SIkm{}.
		(c)~With 108\SIkm{} of total network length, the Lima road network is the longest.
		The depot locations are randomly selected from the vertices of the required edges of the road network.
	The MD-MEM algorithm handles multiple depots within the routing framework.
	\label{fig:large}}
\end{figure*}

We use the multi-depot merge-embed-merge (MD-MEM) to generate solutions.
Figure~\ref{fig:large} shows routes for three road networks.
Due to the large number of edges, it is not possible to solve the problem using the ILP formulation.
Instead, we compare the MD-MEM algorithm with the SRLC-4-approximation~\cite{vanBevernKS17} and the SRLC-3-approximation~\cite{AgarwalA20WAFR} algorithms for the single robot line coverage problem (SRLC).
These two single-robot algorithms are solving a simpler version of the problem as they do not consider capacity constraints and depot locations.
Yet, the MD-MEM algorithm performs significantly better than the single robot algorithms, as shown in \fgref{fig:costs_large}.
In the figure, the cost difference percentage is computed with respect to the total average cost as follows:
\begin{equation*}
	\label{eq:cost_diff}
	\text{cost difference \%} = 100\times\frac{c - \bar c}{\bar c},\  \bar c = \sum_{e\in E_r} \frac{\scost{a_e} + \scost{\bar a_e}}{2}.
\end{equation*}

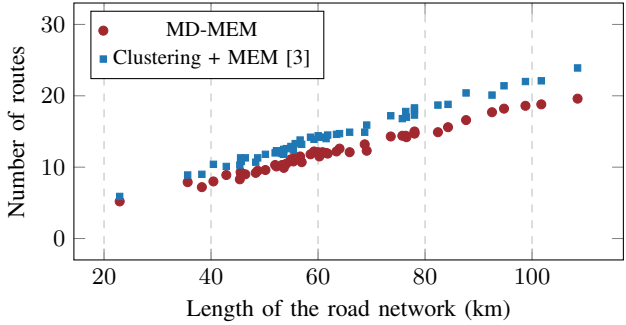
\begin{figure}[t!]
	\centering
	\begin{center}
\begin{tikzpicture}
	\small
	\begin{axis}[
		enlargelimits=true,
		xlabel = {Length of the road network (km)},
		ylabel = {Number of routes},
		xmajorgrids=true,
		grid style=dashed,
		legend pos = north west,
		legend style={font=\footnotesize},
		height=5cm,
		ymin=0,
		ymax=30,
		ylabel near ticks
		]
		\addplot[
		only marks,
		color=mDarkRed,
		mark=*,
		mark size=1.7pt]
		table[x expr= \thisrow{length_m}/1000,y=num_mdmem, col sep=comma]
		{./graphics/data/md_exp_random.csv};
		\addplot+[
		only marks,
		color=mBlue,
		mark options={fill=mBlue},
		mark=square*,
		mark size=1.2pt]
		table[x expr= \thisrow{length_m}/1000,y=num_cluster, col sep=comma]
		{./graphics/data/md_exp_random.csv};
		\legend{MD-MEM, Clustering + MEM~\cite{AgarwalA20ICRA}}
	\end{axis}
\end{tikzpicture}
\end{center}
	\caption{Comparison of the number of routes generated using the clustering approach to generate subgraphs~\cite{AgarwalA20ICRA} and the MD-MEM algorithm.
	The depot locations are selected at random, and the number of routes is the average over 10 runs.
The clustering approach generates subgraphs based on the depot locations and solves each subgraph as a single-depot problem.
The MD-MEM algorithm consistently generates fewer routes than the clustering approach.}
	\label{fig:numroutes}
\end{figure}

In prior work, the multi-depot problem was solved by clustering the edges based on the depot locations as a preprocessing step~\cite{MuyldermansCO03,AgarwalA20ICRA}.
These clusters form a set of subgraphs, and each subgraph is then solved as a single-depot problem.
However, this process yields a larger number of routes, as routes that service edges from different subgraphs are not permitted, thereby restricting the solution space.
In contrast, the MD-MEM algorithm considers the depots within the routing process itself, resulting in a smaller number of efficient routes.
\fgref{fig:numroutes} shows a comparison of the number of routes generated using the clustering approach to create subgraphs and the MD-MEM algorithm---it can be seen that the MD-MEM algorithm consistently generates fewer routes.
As the number of routes corresponds to the number of flights, a reduction in the number of routes can impact the operation costs significantly in terms of equipment and operators required.
In terms of total cost, the MD-MEM consistently performs better than the clustering approach, with an average cost difference of 2.5\%.
The cost difference is not huge as both methods use the same MEM structure to generate routes; the primary advantage is in the reduction of the number of routes.

\begin{figure}[t!]
	\centering
\begin{center}
\begin{tikzpicture}
	\small
	\begin{axis}[
		enlargelimits=true,
		xlabel = {Number of required edges},
		ylabel = {Computation time (s)},
		xmajorgrids=true,
		grid style=dashed,
		legend pos=north west,
		height=5cm,
		ylabel near ticks
		]
		\addplot+[
		color=mDarkRed,
		mark=*,
		mark size=1.7pt,
		mark options={fill=mDarkRed},
		smooth, 
		error bars/.cd, 
		y fixed,
		y dir=both, 
		y explicit]
		table[x=m,y=mean, y error=std, col sep=comma]
		{./graphics/data/mdmem_timing.csv};
	\end{axis}
\end{tikzpicture}
\end{center}
	\vspace{-0.4\baselineskip}
	\caption{Computation time of the MD-MEM algorithm for very large graphs, averaged over 100 runs.
	For the largest graph with 4831 required edges, the algorithm can generate solutions within 100\SIs{}.%
	\label{fig:mdmem_time}}
\end{figure}
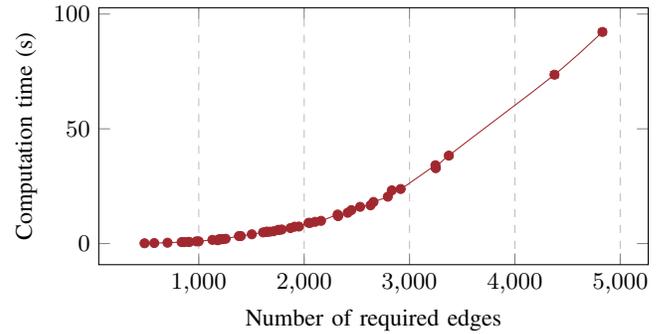

\fgref{fig:mdmem_time} shows the computation time for the MD-MEM algorithm on the large road networks dataset.
The computation time is averaged over 100 runs.
Only the time required by the MD-MEM algorithm is considered, and the time for loading the dataset is not included.
The standard deviation is very low (less than 2\SIs{}) and is not visible in the figure.
The computation time for the largest network consisting of 4831 required edges is within 100\SIs{}.
These experiments establish the reliability of the MD-MEM algorithm to generate high-quality solutions with low computation effort for large line coverage networks.

\subsection{UAV Experiment on a Road Network with a Single Depot}
\begin{figure*}[htbp]
	\centering
	\subfloat[Input Road Network]{%
	\includegraphics[height=0.20\textheight,trim={3cm 0cm 4cm 0cm},clip]{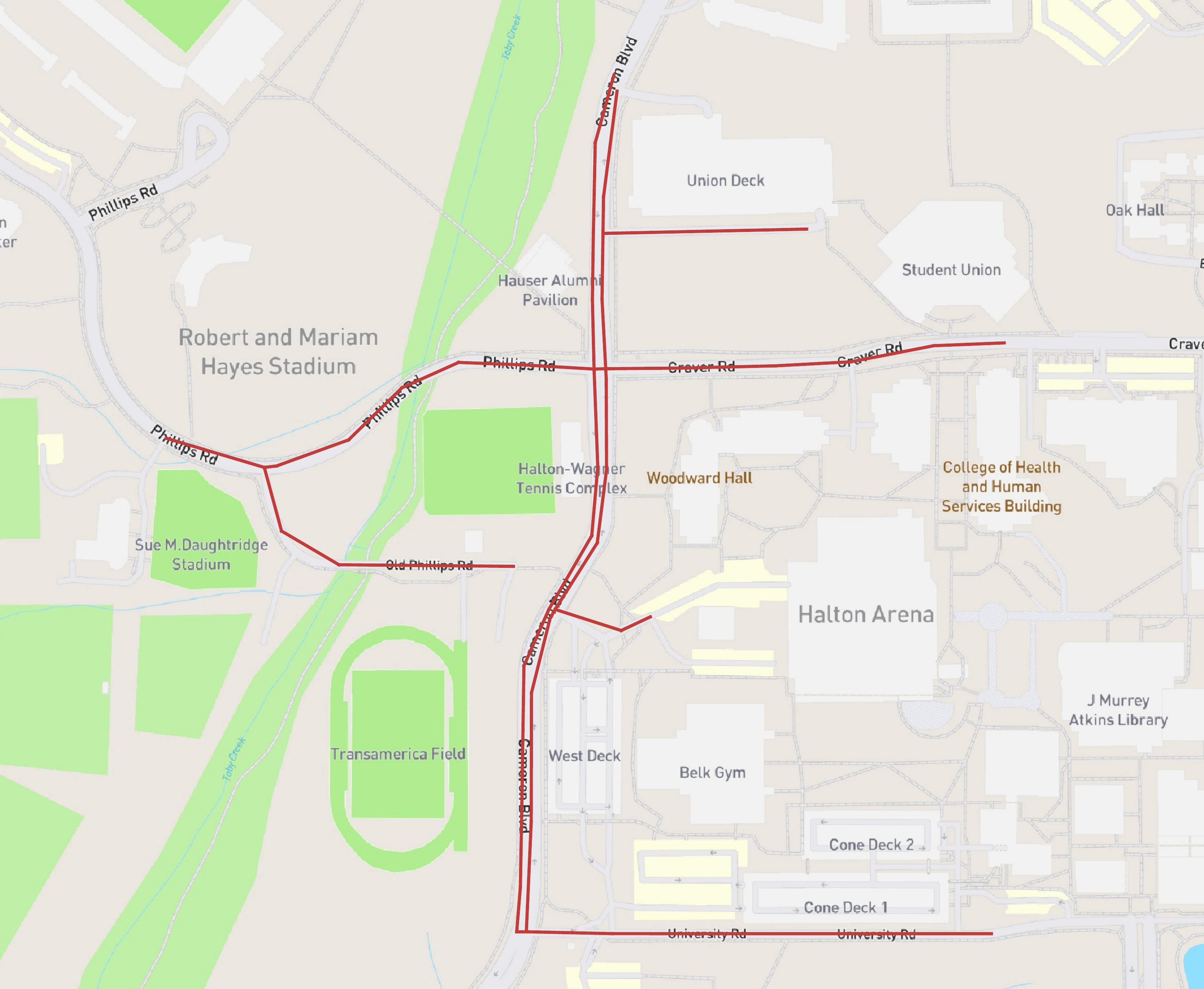}}
	\hfill
	\subfloat[Line Coverage Routes]{%
	\includegraphics[height=0.20\textheight]{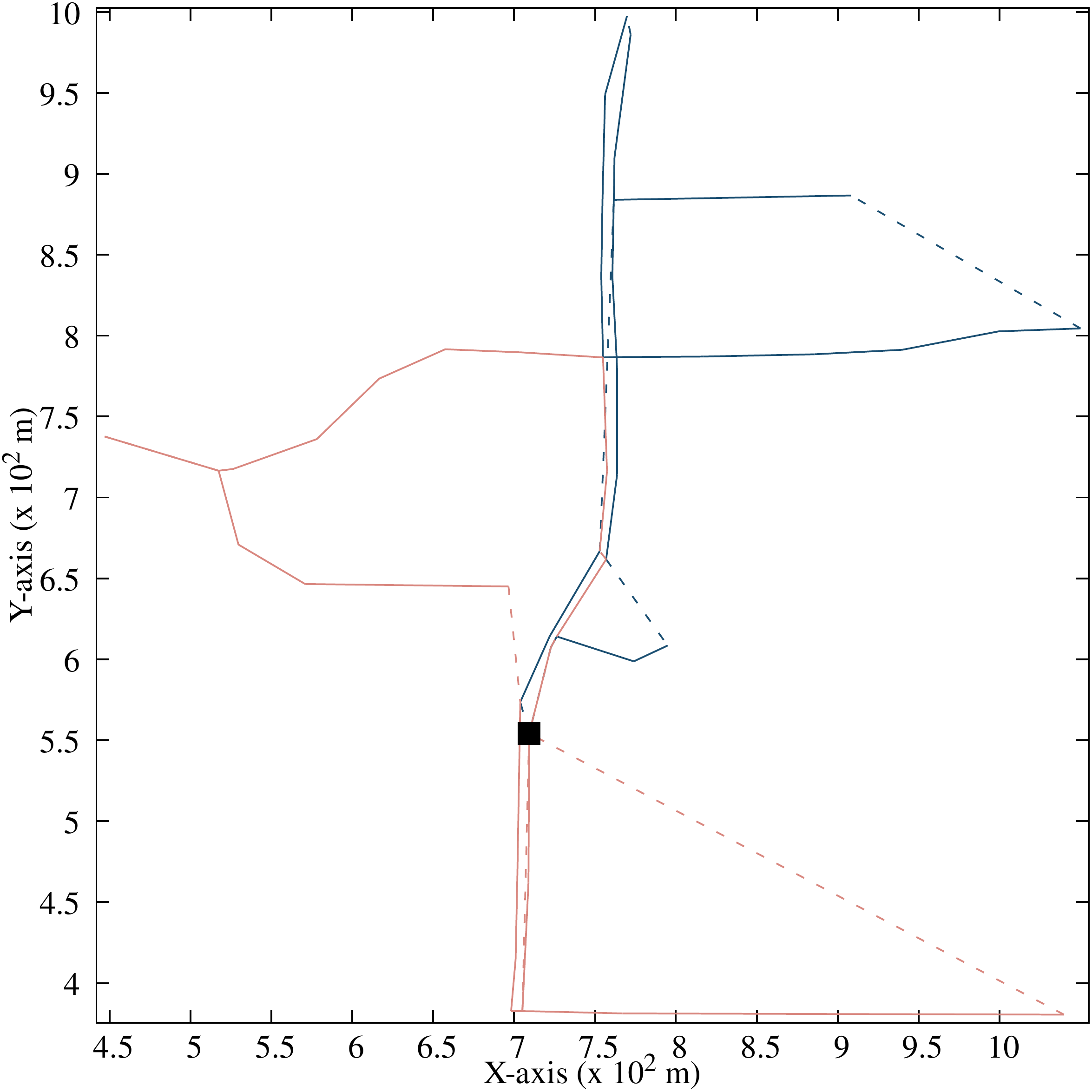}}
	\hfill
	\subfloat[Orthomosaic]{%
	\includegraphics[height=0.20\textheight]{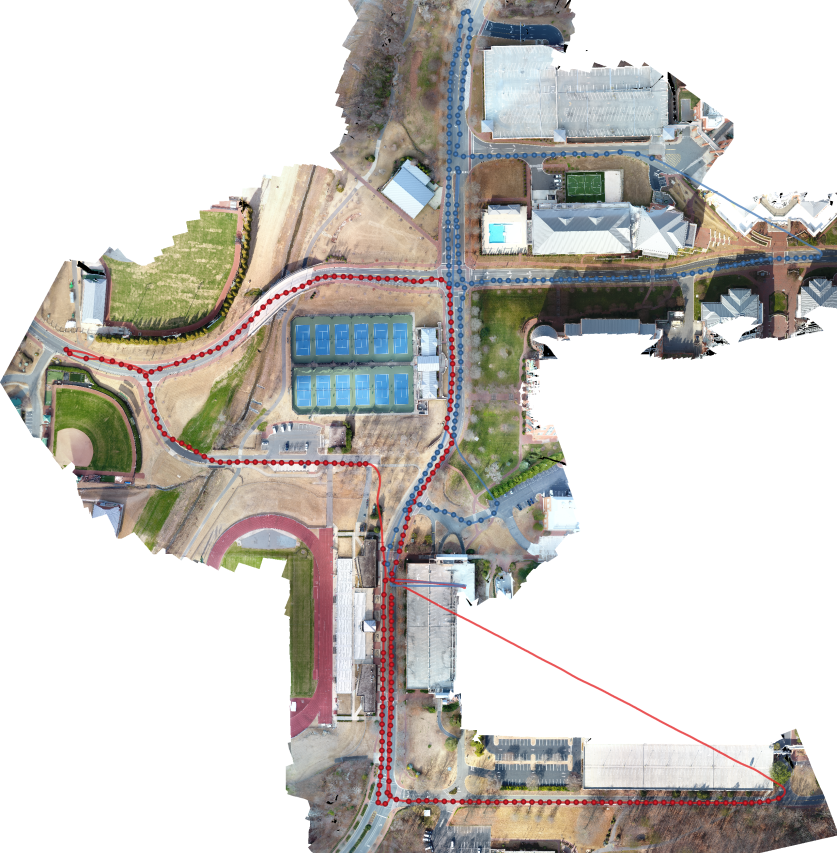}}
	\caption[Line coverage of a portion of the UNC Charlotte road network with two routes and a single depot]{Line coverage of a portion of the UNC Charlotte road network.
		(a)~The input road network has a length of 2,658\SIm{} with 48 vertices and 48 required edges (shown as red lines).
		There are 1,128 non-required edges formed by pairs of vertices (not shown).
		(b)~Two routes distinguished by different colors are computed using the MEM algorithm.
		The algorithm computes a depot location, shown by the black square, from where the UAVs start and end their routes.
		The solid lines represent servicing, while the dashed lines represent deadheading.
		(c)~An orthomosaic generated from the images taken by the UAVs flown autonomously along the computed routes.
		The lines show the actual flight path, and the dots are the locations where the images were taken.
	\label{fig:uncc_cl2}}
\end{figure*}

We performed experiments with a UAV on a portion of the UNC Charlotte campus road network, shown in \fgref{fig:uncc_cl2}.
The service and the deadhead speed were set to 3.33\SIvel{} and 5\SIvel{}, and a wind speed of 0.89\SIvel{} from the west was incorporated based on the wind conditions on the day of the experiment.
A conservative capacity of 600\,s was selected.
Coverage routes were generated using the ILP formulation and the MEM algorithm.
The ILP formulation gave the optimal solutions.
The computed and the actual flight times are shown in Table~\ref{tb:uncc_cl2}.
The actual flight times include take-off and landing and are close to the computed cost.
\begin{table}[t!]
	\renewcommand{\arraystretch}{1.5}
	\centering
	\caption{Comparison of computed and actual flight times}
	\label{tb:uncc_cl2}
	\begin{tabular}{@{}ccccc@{}}
		\toprule
 &
		\multicolumn{2}{c}{\begin{tabular}[c]{@{}c@{}}Computed Cost (s)\end{tabular}} &
		\multicolumn{2}{c}{\begin{tabular}[c]{@{}c@{}}Actual Flight Time (s)\end{tabular}} \\ \cmidrule(l){2-3} \cmidrule(l){4-5} 
																																									& Route 1 & Route 2 & Route 1 & Route 2 \\ \cmidrule(l){2-5}
		ILP & 462     & 554     & 527     & 642     \\
		MEM & 473     & 599     & 548     & 688    \\
		\bottomrule
	\end{tabular}
\end{table}

The experiment shows that the line coverage problem is well-suited to model coverage of linear infrastructure such as a road network.
The problem models two different modes of travel and incorporates wind conditions in the formulation.
Furthermore, activating the sensors only during servicing helps in reducing the amount of data that requires post-processing.

\subsection{Experiment on a Large-Scale Road Network}
We used the multi-depot MEM algorithm for the UNC Charlotte campus road network spanning an area of 1.5\SIkmsqr{} and a length of 12\SIkm{}, shown in \fgref{fig:uncc}.
The road network consists of 842 vertices, 865 required edges, and 353,196 non-required edges.
$k$-medoids clustering was used to obtain eight depot locations, shown as black squares in the figure.
The depots are well distributed in the road network such that no road segment is far away from all the depots.
The multi-depot MEM algorithm computed eight routes, and no two routes share the same depot.
Note that the $k$-medoids algorithm is only used to compute the depot locations, and we do not cluster the edges.
Service and deadhead speeds of 5\SIvel{} and 8\SIvel{} were set for the experiments.
The computed routes and the orthomosaic generated from the images collected during flights are shown in \fgref{fig:uncc}.
\tbref{tb:fl} gives computed costs, actual flight times, and the number of images collected for each route.
For these experiments, we did not incorporate the wind conditions, which could be a reason for a higher deviation between the computed cost and the actual flight time.

It can be observed from the routes that the individual routes primarily consist of road segments that are close to each other and are connected, indicating that the MEM algorithm can efficiently distribute the line features among routes.
Furthermore, the routes are assigned to the closest depot, showing that the multiple depot formulation of the MEM algorithm is well-suited for the line coverage problem with large graphs.

\begin{table}[t]
	\centering
	\caption{Data of flights for the UNC Charlotte Road Network}
	\label{tb:fl}
	\begin{tabular}{rrr}
		\toprule
		Computed Cost (s) & Flight time (s) & Number of images\\\midrule
		215 & 346 & 87\\
		458 & 616 & 180\\
		407 & 496 & 170\\
		538 & 657 & 221\\
		371 & 493 & 140\\
		449 & 599 & 192\\
		481 & 627 & 210\\
		277 & 393 & 120\\
		\bottomrule
	\end{tabular}
\end{table}
\subsection{Nonholonomic Robots}
\begin{figure}[htbp]
	\begin{center}
\begin{tikzpicture}
	\small
	\begin{axis}[
		enlargelimits=true,
		xlabel = {Length of the road network (km)},
		ylabel = {Cost difference (\%)},
		xmajorgrids=true,
		grid style=dashed,
		legend pos = north east,
		legend style={font=\footnotesize},
		height=5.5cm,
		ylabel near ticks
		]
		\addplot[
		only marks,
		color=mDarkRed,
		mark=*,
		mark size=1.7pt]
		table[x expr= \thisrow{L}/1000,y=memDtwo, col sep=comma]
		{./graphics/data/nonholonomic.csv};
		\addplot+[
		only marks,
		color=mBlue,
		mark options={fill=mBlue},
		mark=square*,
		mark size=1.2pt]
		table[x expr= \thisrow{L}/1000,y=memD, col sep=comma]
		{./graphics/data/nonholonomic.csv};
		\addplot+[
		only marks,
		color=mBlue,
		mark options={fill=mGreen},
		mark=diamond*,
		mark size=2.0pt]
		table[x expr= \thisrow{L}/1000,y=post, col sep=comma]
		{./graphics/data/nonholonomic.csv};
		\legend{MEM-Turns, MEM-Turns-Dubins, MEM + Post-Process}
	\end{axis}
\end{tikzpicture}
\end{center}
	\caption{%
		Performance analysis of the MEM algorithm for nonholonomic robots on the smaller road network dataset with a single depot.
		The cost difference is measured with respect to MEM	solutions without nonholonomic constraints.
		The green diamonds correspond to the post-processing of holonomic routes to satisfy the nonholonomic constraints.
		The MD-MEM-Turns algorithm is executed with only Dubins curves (blue squares) and with both Dubins curves and smooth turns (red circles).
	Incorporating both Dubins curves and smooth turns in the MEM algorithm consistently computes lowest cost solutions.}
	\label{fig:nonholonomic}
\end{figure}
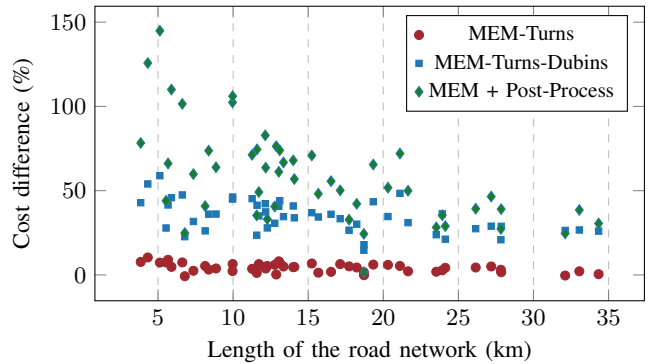
We analyzed the performance of the MD-MEM-Turns algorithm for nonholonomic robots on the smaller 50 road network dataset with a single depot.
The service and the deadhead speeds were set to 3.33\SIvel{} and 5.00\SIvel{}.
The maximum angular velocity and acceleration were set to $\pi/4$\SIomega{} and 3.00\SIacc{}, and the flight time was limited to 1800\SIs{}.
A deviation limit $\delta_{\max}$ of 2\,m was selected to allow smooth turns for adjacent required edges.
The results, shown in \fgref{fig:nonholonomic}, indicate that incorporating smooth turns (red circles) consistently computes solutions with significantly lower costs than just using Dubins curves (blue squares).
The post-processing approach (green diamonds) does not perform as well as the MD-MEM-Turns algorithm, as it does not consider the turning costs during the route generation process.
Furthermore, it can also violate the capacity constraints.
When using only Dubins curves with the MEM algorithm, the average improvement over the post-processing approach is 13\%, whereas using both Dubins curves and smooth turns gives an improvement of 32\%.

\begin{figure}[htpb]
	\centering
	\subfloat[Input Road Network]{%
	\includegraphics[height=0.18\textheight]{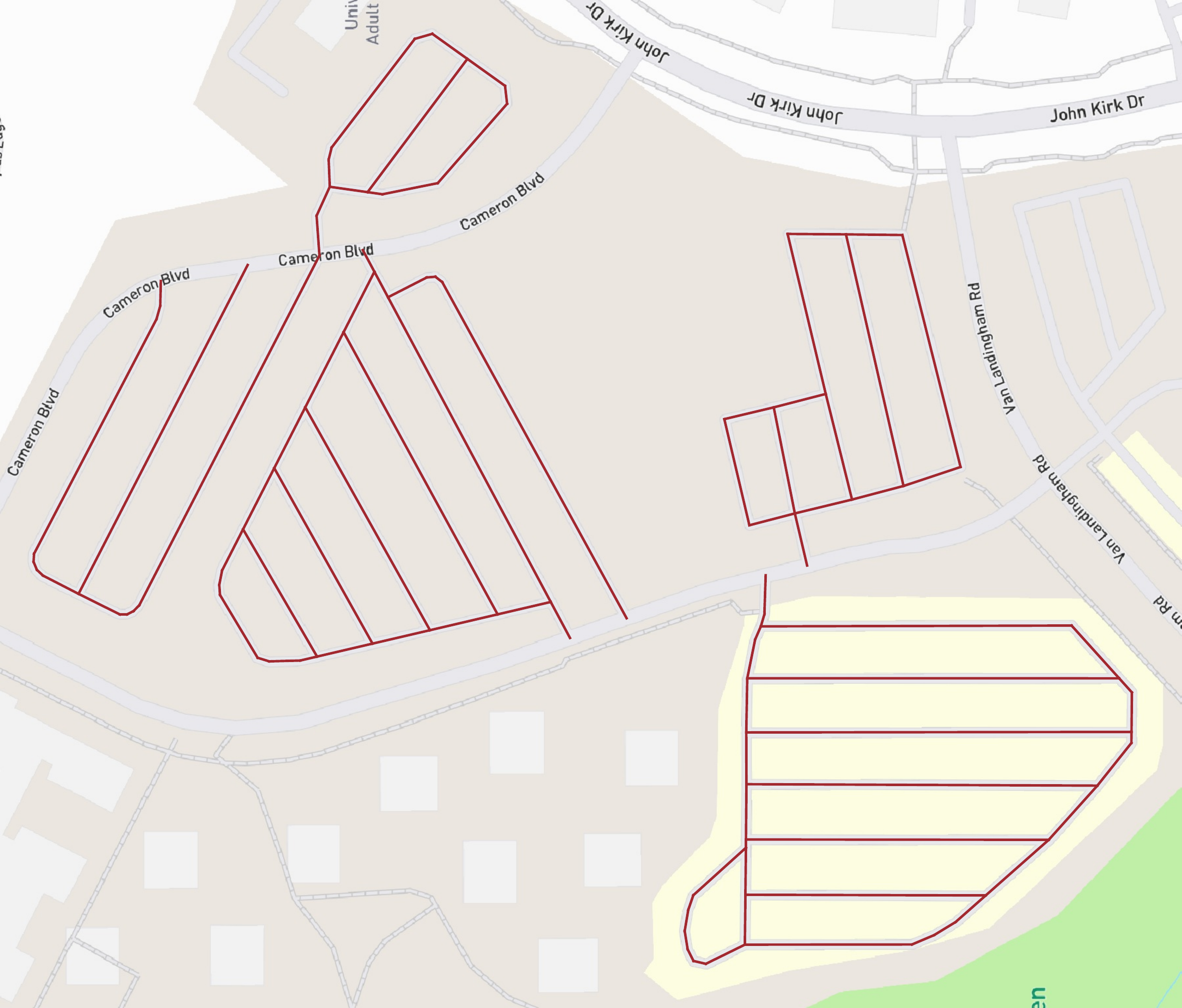}}
	\hspace{1cm}
	\subfloat[Line Coverage Routes]{%
	\includegraphics[height=0.18\textheight]{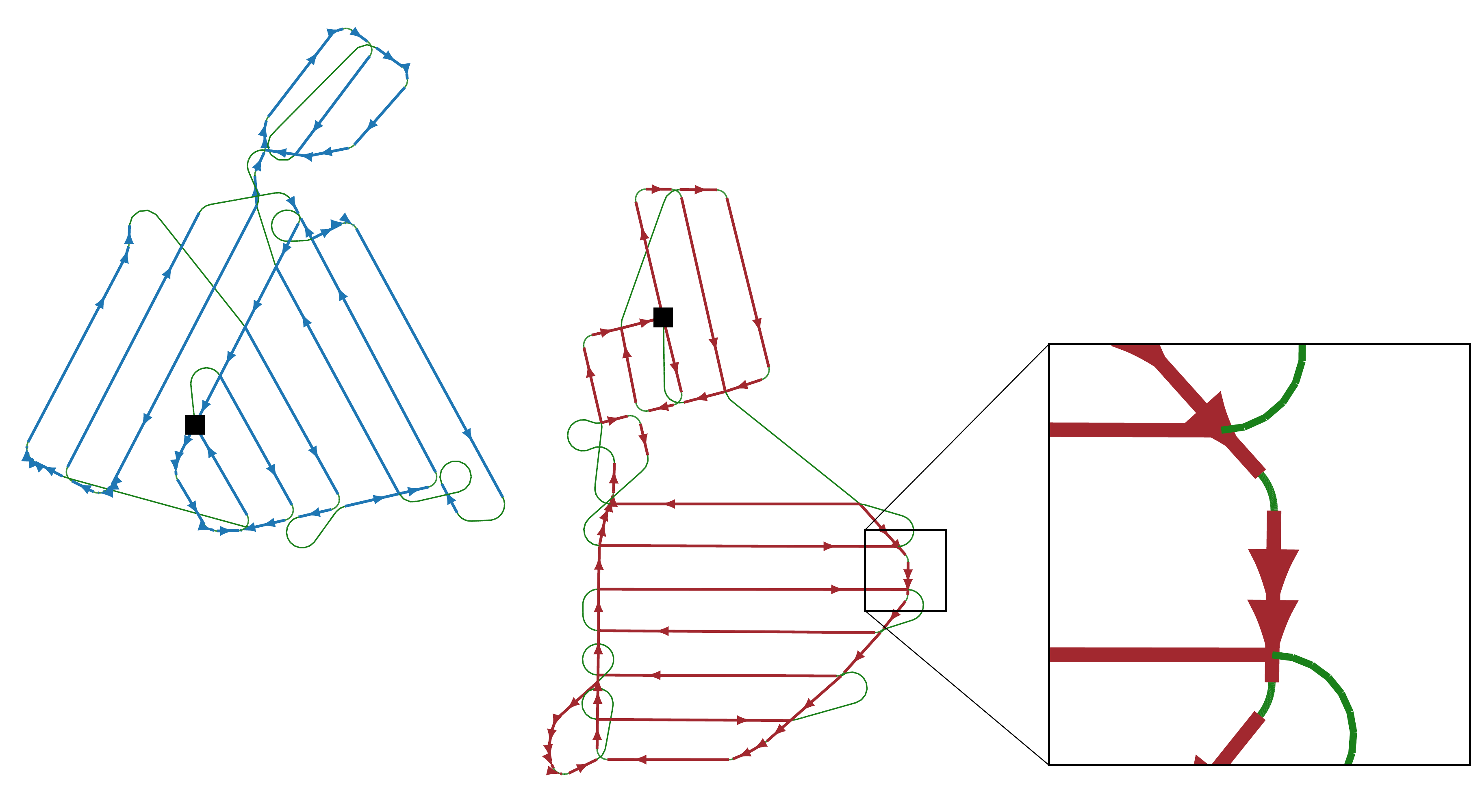}}
	\caption[Line coverage of a network of lanes in a set of parking lots using nonholonomic robots and multiple depots]{Line coverage of a network of lanes in a set of parking lots:
		(a)~The total length of the lanes in the input graph is 2,982\SIm{}.
		There are 90 vertices, 104 required edges, and 4,005 non-required edges.
		(b)~Coverage routes using two depots and nonholonomic robots.
		There are two routes distinguished by different colors.
		The green lines show deadheadings composed of Dubins curves and smooth turns.
		The arrows indicate the direction of travel.
	The inset shows an enlarged view of smooth turns for adjacent required edges.%
	\label{fig:lot56}}
\end{figure}

\fgref{fig:lot56} shows a network of lanes on a set of parking lots as an illustrative example.
Two routes that respect the nonholonomic constraints are generated using the MD-MEM-Turns algorithm with two depots and a flight time limit of 600\SIs{}.
The deadheadings are composed of Dubins curves and smooth turns for non-adjacent and adjacent required edges, respectively.
Note that the algorithm computed routes that cover separated regions while reducing the amount of deadheading, thereby optimizing the total cost of the routes.
The computation time of the MD-MEM-turns algorithm is similar to that of the standard MEM algorithm because of the constant time computation of savings, as discussed in \scref{sec:turns}.

\begin{figure*}[t]
	\centering
	\subfloat[Cell decomposition]{%
	\includegraphics[height=0.18\textheight]{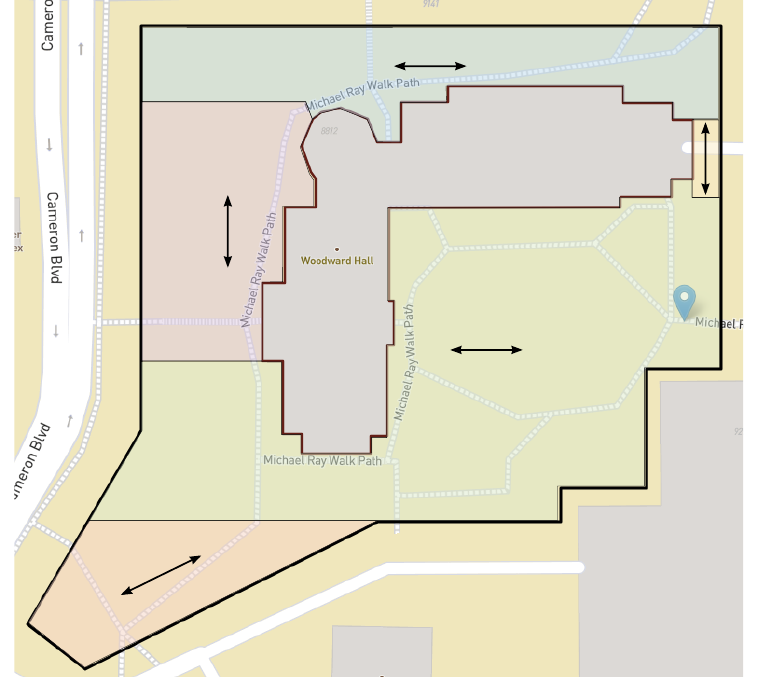}}
	\hfill
	\subfloat[Service tracks]{%
	\includegraphics[height=0.18\textheight]{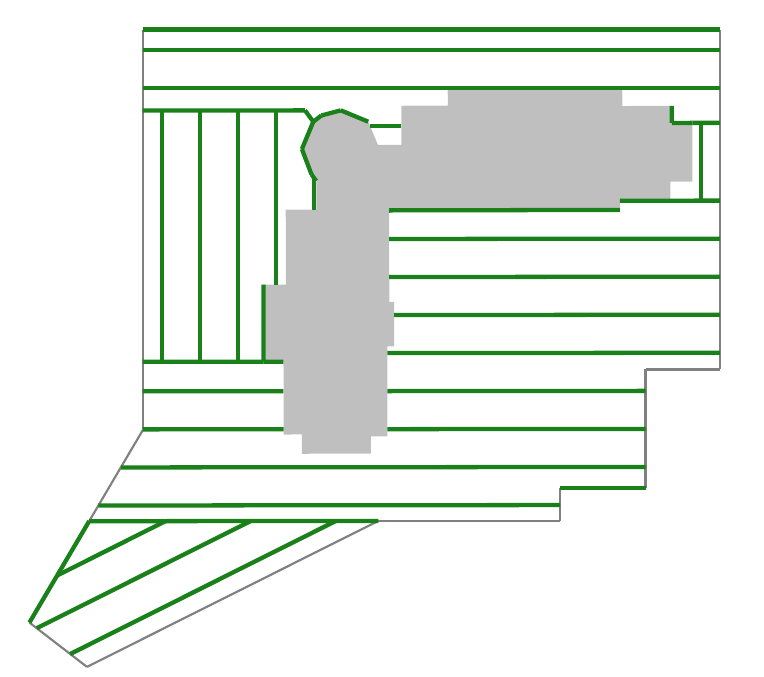}}
	\hfill
	\subfloat[Routes with nonholonomic robots]{%
	\includegraphics[height=0.18\textheight]{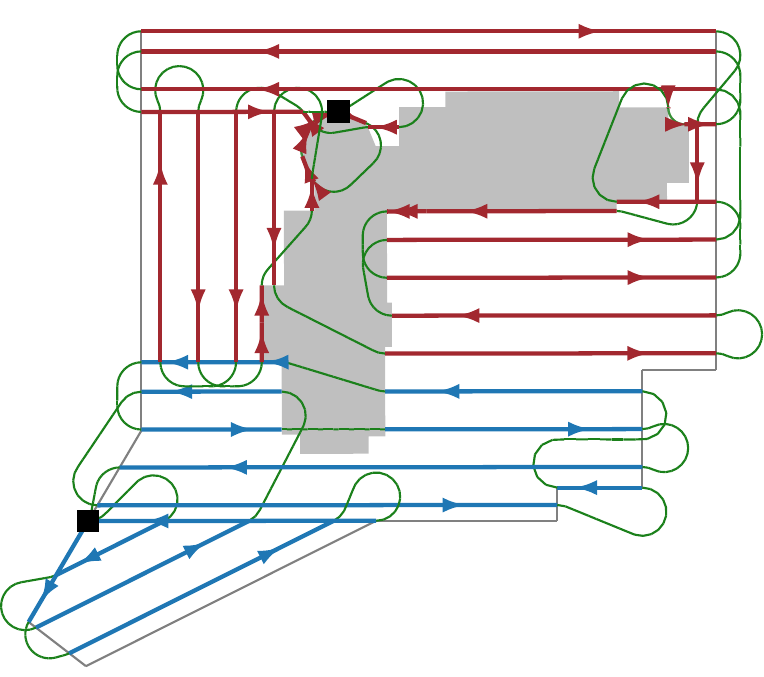}}
	\caption{%
Area coverage using aerial robots for a real-world example from~\cite{AgarwalA22RAL}: (a)~The region surrounding a building needs to be covered. The cell decomposition is shown with double-head arrows indicating service directions.
(b)~Service tracks are generated for each cell independently. The tracks are parallel to the service directions obtained from cell decomposition. These tracks form the linear features for the line coverage problem.
(c)~Two routes computed by the MEM algorithm for multiple depots and nonholonomic robots. The green lines indicate deadheading and comprise Dubins curves between non-adjacent tracks and smooth turns between adjacent tracks.
	\label{fig:ww}}
\end{figure*}

\subsection{Area Coverage: Multiple Depots and Nonholonomic Robots}
We now illustrate the use of the algorithms developed for the line coverage problem to generate routes for area coverage.
Given a region $\mathcal R \subset\mathbb R^2$, the area coverage problem is to find a set of routes for a team of robots such that the total cost of the routes is minimized, and the robots service all the points in the region.
The sensor field-of-view is an important parameter that needs to be factored in when determining the routes.
We developed a procedure to transform the area coverage problem to the line coverage problem and then used the MEM algorithm to generate efficient routes that consider resource constraints and asymmetric costs~\cite{AgarwalA22RAL}.

The formulation for solving the area coverage problem consists primarily of three components:
(1)~{\em Cell decomposition} of the environment,
(2)~{\em Service track generation} for individual cells, and
(3)~{\em Routing} to traverse the service tracks.
The service tracks form the linear features that robots must service.
The problem is formulated as a graph, and efficient line coverage algorithms are then used to generate routes that traverse the service tracks.
Moreover, the transformation facilitates a significant generalization of the cell decomposition component to reduce the number of turns the robots must take.
In particular, the cells are no longer required to be {\em monotone} polygons~\cite{deberg} with respect to the service direction.
This generalization enables additional service directions for the cells to minimize the number of turns and reduces the number of service tracks by avoiding overlapping sensor coverage regions at the common boundary of adjacent cells.
It was established in~\cite{AgarwalA22RAL} that the above approach, aided by the line coverage algorithms, gives solutions that are better than other approaches in the literature.

The line coverage algorithms for multiple depots and nonholonomic constraints extend to the area coverage problem.
\fgref{fig:ww} illustrates solutions computed using the MD-MEM-Turns algorithm in an outdoor environment setting from~\cite{AgarwalA22RAL}.
The environment spans a 19,000\SImsqr{} region with a building containing 45 vertices.
As UAVs can fly at high altitudes, we allow non-required edges that cross the building.
However, only the region surrounding the building needs to be covered.
\fgref{fig:ww}(a) shows the cell decomposition of the region into five cells.
The service tracks, shown in \fgref{fig:ww}(b), take the field-of-view of the sensor into account.
These service tracks form the linear features, i.e., the required edges, for the line coverage problem.
Dubins curves are used to deadhead between pairs of non-adjacent required edges, and smooth turns are used to deadhead between adjacent required edges.
\fgref{fig:ww}(c) shows the two routes computed using the MEM algorithm with multiple depots and nonholonomic constraints.
The example illustrates the use of line coverage algorithms for area coverage, and shows that the enhancements to algorithms for line coverage translate directly to area coverage.

\section{Conclusion}
\label{sc:conclusion}
%
The paper presented the line coverage problem with multiple resource-constrained robots for coverage of linear features such as road networks and power lines.
The environment was modeled as a graph with the linear features as a set of required edges.
The formulation permits non-required edges that can enable faster travel, but the robots need not cover them.
The edges have non-negative costs (e.g., travel time) and resource demands (e.g., battery life), which can be direction-dependent to facilitate the modeling of wind conditions, uneven terrain, and one-way streets.
The goal is to compute routes that together cover all the required edges while minimizing the total cost of the routes under the constraint that each route's total demand is within the robot's resource capacity.


We posed line coverage as an optimization problem on graphs and formulated it as an integer linear program (ILP).
As the problem is NP-hard, we designed a heuristic algorithm, Merge-Embed-Merge (MEM), with a time complexity of $\mathcal O(m^2\log m)$, where $m$ is the number of linear features.
Leveraging the constructive nature of the MEM algorithm, we extended the algorithm to multiple variants of the line coverage problem.
To address large graphs, we formulated the line coverage problem with multiple depots and designed the Multi-Depot MEM (MD-MEM) algorithm.
We incorporated smooth turns and nonholonomic constraints, resulting in the MD-MEM-Turns algorithm.
These are the first set of polynomial-time algorithms that directly incorporate multiple depots, smooth turns, and nonholonomic constraints into the line coverage problem with resource-constrained robots.

The algorithms were evaluated on medium and large road networks from the world's 50 most populous cities, which showed that the MEM algorithm computes high-quality solutions with costs within 7\% of the ILP solutions.
The MEM algorithm also performs similarly when the capacity of the robots is varied.
The computation times show that the algorithm is very fast, solving single-depot road networks within a fraction of a second.
On large road networks, the MD-MEM algorithm generates solutions that are significantly better than the existing approximation algorithms for the single robot line coverage problem.
When compared to existing clustering-based approaches, the MD-MEM algorithm consistently generates fewer routes, thereby optimizing operating costs.

The algorithms were demonstrated in physical experiments on the UNC Charlotte road network.
For the first experiment, two routes from a single depot location were autonomously executed by a commercial UAV to cover a portion of the road network.
In the second experiment, the UAVs executed routes from the multiple depot formulation for a large road network.

We showed that the new capabilities of incorporating multiple depots and nonholonomic robots translate directly to routing for the area coverage problem.
It is easy to transform point features to degenerate linear features by creating artificial edges.
Thus, all three types of features---points, lines or curves, and areas---can be modeled within the same {\em generalized coverage}~\cite{AgarwalThesis} framework.

Future work includes addressing line coverage with a heterogeneous team of robots that vary in their resource capacity and sensor capabilities.
We are also interested in exploring the orienteering routing problem in the context of linear features, where the features have associated profits, and the objective is to service edges that maximize the total profit.
Since the MEM algorithm is very fast and can potentially be used for computing routes online on the robots, we plan to characterize the performance of the algorithm in distributed and online settings.
From a theoretical standpoint, establishing an approximation guarantee or proving the inapproximability of the problem is an important direction for future work.

\section*{Acknowledgments}
\noindent The authors thank Mr. Rick Boucher, Jr. from Parking and Transportation Services at UNC Charlotte for helping us with the UAV experiments. Thanks to the anonymous reviewers for their comments that have helped strengthen the paper.

\bibliographystyle{IEEEtran}

\end{document}